# Guided interactive image segmentation using machine learning and color-based image set clustering

**Adrian Friebel[1], Tim Johann[2], Dirk Drasdo[2,3], Stefan Hoehme[1*]**

1 Institute of Computer Science, Leipzig University, Haertelstr. 16–18, 04107, Leipzig, Germany; 2 IfADo - Leibniz Research Centre for Working Environment and Human Factors, Ardeystrasse 67, Dortmund, Germany; 3 INRIA Paris & Sorbonne Université LJLL, 2 Rue Simone IFF, 75012, Paris, France; *Corresponding author

**Over the last decades, image processing and analysis has become one of the key technologies in systems biology and medicine. The quantification of anatomical structures and dynamic processes in living systems is essential for understanding the complex underlying mechanisms and allows, i.a., the construction of spatio-temporal models that illuminate the interplay between architecture and function. Recently, deep learning significantly improved the performance of traditional image analysis in cases where imaging techniques provide large amounts of data. However, if only few images are available or qualified annotations are expensive to produce, the applicability of deep learning is still limited.**

**We present a novel approach that combines machine learning based interactive image segmentation using supervoxels with a clustering method for the automated identification of similarly colored images in large image sets which enables a guided reuse of interactively trained classifiers. Our approach solves the problem of deteriorated segmentation and quantification accuracy when reusing trained classifiers which is due to significant color variability prevalent and often unavoidable in biological and medical images. This increase in efficiency improves the suitability of interactive segmentation for larger image sets, enabling efficient quantification or the rapid generation of training data for deep learning with minimal effort. The presented methods are applicable for almost any image type and represent a useful tool for image analysis tasks in general.**

**The provided free software TiQuant makes the presented methods easily and readily usable and can be downloaded at tiquant.hoehme.com.**

# Introduction

Advances in imaging technology have led to a diversification of microscopy techniques enabling examination of a broad spectrum of biological questions and have thus contributed to the establishment of imaging and image analysis as one of the main pillars of bioscience. Optical microscopy in combination with (immuno)histochemical and immunofluorescent staining allows for visualization of tissue and physiological entities. Therefore, these techniques are widely used (1) in medical diagnosis, e.g., using histological sections of tissue specimens or biopsies, as a means to distinguish physiological and pathological tissue architectures, and (2) in medical research to study tissue microarchitecture, e.g., using 3D volumetric confocal microscopy [1–3], and tissue-scale or intracellular processes, e.g., by time resolved two-photon microscopy [4]. In many use-cases, computer-aided image analysis is preceded by detection or pixel-accurate segmentation of objects of interest. Detection asks for the location of an object in the image, e.g., using bounding boxes, while segmentation tries to capture the precise shape of the object. The most basic segmentation technique, manual pixel-accurate labeling, is time consuming, inconsistent and practically infeasible in many use cases, such as e.g., blood vessel segmentation in volumetric images. A multitude of (semi-)automatic image processing methods have been proposed, including e.g., intensity-thresholding and morphological operators [1–3], region-based methods [5] or deformable models [6]. These methods are usually tailored towards a specific image setup (i.a., image dimensionality, magnification, staining) and/or object class (e.g., nuclei, blood vessels, necrotic tissue) and as such are typically limited in their applicability to differing use cases. Additionally, method parametrization to cope with image variability is often challenging and time consuming. More recently, shallow and deep machine learning methods were successfully applied to pixel-accurate image segmentation [7–10]. By learning object appearance from training examples these methods allow for incorporation of subtle expert knowledge, are less restricted regarding image or object class specifics and minimize parametrization complexity.

Deep learning proved to be an extremely powerful approach to improving performance across all image processing tasks [9, 11–13]. This is primarily due to its ability to operate directly on the input image and implicitly learn features of increasing complexity. The price of finding a suitable feature space in addition to the decision surface itself is a high demand for training data and long training times. By now there is a large variety of specialized network architectures for segmentation of biomedical images and pre-trained models for standard targets such as nuclei or cells that can be applied directly if the images to be segmented are similar to images in the training set, or that can be tuned using transfer learning thereby reducing the amount of training data needed. However, the range of imaging systems and segmentation targets in the biomedical field is diverse so that often no suitable pre-trained models are available, image counts are in many cases very low and ground truth is usually not readily available and expensive to generate, which poses significant challenges for the applicability of deep learning methods. Furthermore, the development but also the training of deep learning models requires significant expert knowledge and is therefore difficult to achieve for untrained personnel. In contrast, traditional, so called shallow, machine learning approaches such as random forests or support-vector machines, rely on handcrafted features. While this limits versatility, as engineered features need to represent the learning problem at hand, finding a decision surface in a restricted feature space greatly simplifies the learning task. Thereby, relatively few training examples and short training times are required, which makes these shallow methods eminently suitable for interactive segmentation.

A number of interactive segmentation approaches and software packages that utilize machine learning have been published in recent years. A supervoxel-based approach for segmentation of mitochondria in volumetric electron microscopy images was developed by [7], though the published executable software and code encompassed only the supervoxel algorithm *SLIC*. SuRVos [14] is a software for interactive segmentation of 3D images using a supervoxel-based hierarchy, which lays a focus on segmentation of noisy, low-contrast electron microscopy datasets. The extendable software package Microscopy Image Browser [15] allows for processing of multidimensional datasets and features a selection of conventional processing algorithms, several region selection methods aiding manual segmentation, and methods for semi-automatic segmentation, including a supervoxel-based classification approach. FastER [16] is designed specifically for cell segmentation in grayscale images, using features that are very fast to compute. The Fiji plugin Trainable Weka Segmentation [17] is a tool for pixel classification in 2D and 3D images using a broad range of features. Ilastik [18] provides a streamlined user interface with workflows for i.a., image segmentation, object classification and tracking as well as a sophisticated on-demand back end enabling processing of images with up to five dimensions that are larger than available RAM. Its segmentation workflow employs a random forest classifier for pixel classification from local features such as color, edge-ness and texture at different scales.

Most of these interactive segmentation tools allow for reuse of trained classifiers on new, unseen images. However, due to the variability of image appearance, even for sets of images that were acquired following a standard protocol, classifier reuse is usually a trial-and-error procedure and quickly becomes cumbersome for larger image sets. One of the main contributing factors to image quality variability are systematic discrepancies of coloration that can be attributed to minor deviations in the image acquisition process (e.g., sample preparation procedure, imaging settings, condition of the imaged subject). Since color information is used in various ways as a feature by all mentioned segmentation tools that target colored images, degraded prediction accuracy is to be expected for images with variable coloration.

We propose an approach that combines *(i)* interactive image segmentation from sparse annotations with *(ii)* the guided reuse of thereby trained classifiers on unseen images to enable efficient batch processing. *(i)* We formulate interactive pixel-accurate segmentation as a machine learning problem working on supervoxels, which are connected, homogeneously colored groups of voxels, using random forests or support vector machines as classifiers. Dimensionality reduction through use of supervoxels, precomputation of supervoxel features and a convenient graphical user interface enable rapid, intuitive refinement of training annotations by iterative correction of classification errors or uncertainties. *(ii)* We introduce a color-based image clustering method that enables automated grouping of image sets into subsets of similarly colored images. A corresponding number of so-called prototype images is identified which serve as eligible candidates for interactive training of classifiers for within-subset reuse. These new methods, that are applicable to two- and three-dimensional images, have been fully incorporated into our freely available image processing software TiQuant. The first version of TiQuant, released in 2014, included various processing methods specifically for liver tissue segmentation of 3D confocal micrographs, as well as the corresponding analysis functionality [1, 2].

We evaluate the interactive image segmentation method as well as the color-based image clustering strategy to guide classifier reuse using a previously published image set consisting of 22 brightfield micrographs of mouse liver tissue with corresponding manual nuclei annotations [3]. We show that the interactive approach outperforms a human annotator and a comparable state-of-the-art interactive segmentation software and that limiting classifier reuse to similarly colored images significantly improves performance compared to reuse on images of differing coloration, yielding results close to the level of a human annotator.

# Method

The general workflow of segmenting an image with our supervoxel-based approach is split into a preprocessing step and the interactive training, prediction and segmentation steps, that require direct user intervention and must be iteratively repeated for refinement as needed to achieve a segmentation of desired quality (see *Fig.1A*).

In the preprocessing step, the image is partitioned into similarly sized supervoxels and descriptive features are computed for them. A single parameter, the target supervoxel size, must be adjusted by the user to ensure supervoxels fit the objects of interest. This step is computationally expensive compared to the interactive steps because it scales with the number of pixels and superpixels.

For example, on a modern computer, it takes ~1 minute to partition a 7.5k x 7.5k pixel image into 100k superpixels and six minutes to calculate all available features for them. However, this step is usually performed only once per image.

Subsequently, the user annotates exemplary fore- and background regions in the original image to generate a training database. A classifier is then fitted to the training data, and probability estimates for class membership are predicted for all supervoxels in the image. In the final step, a segmentation is generated based on these probability estimates, which can be refined through post-processing.

If segmentation quality is insufficient, the user can provide further training data, especially for regions that were poorly segmented. Thereby, high-quality segmentations can be generated quickly by iterative annotation refinement.

Trained classifiers can be used for prediction on unseen images, eliminating the need for producing training data for every image. To guide the process of identifying training images and matching candidate images for classifier reuse, the images to be processed are grouped into similarly colored subsets using a two-stage clustering approach (see *Fig.1B*).

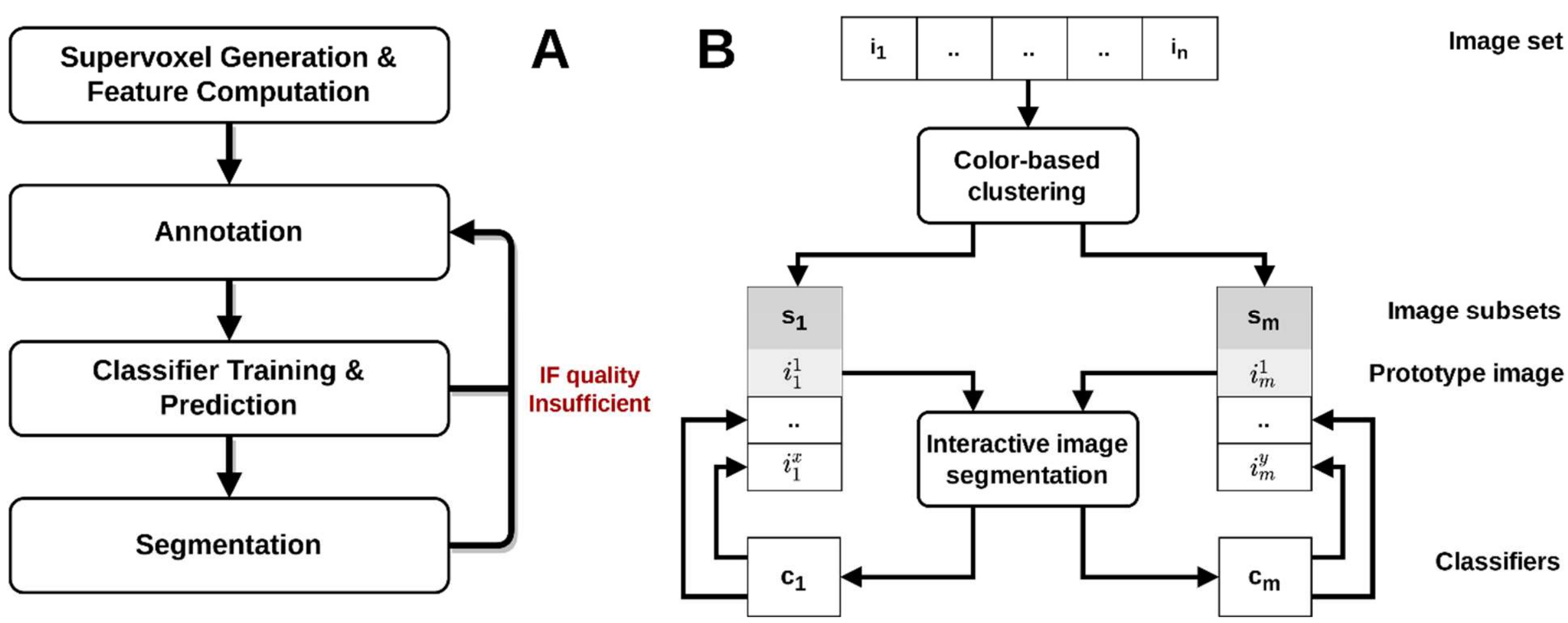

**Fig.1: (A)** Workflow of the supervoxel-based interactive image segmentation approach from a user perspective. The classifier training procedure is illustrated in detail in *Fig. 3A*. **(B)** Workflow of guided reuse of interactively trained classifiers: The set of *n* images is grouped into *m* subsets of similarly colored images using our color-based image clustering method. For each subset a prototype image is identified, which is then interactively segmented. The resulting *m* trained classifiers are reused on the remaining images in their respective subsets. The color-based clustering algorithm is depicted in detail in *Fig. 3B*.

## Supervoxel initialization

The term supervoxel (respectively superpixel in 2D) was introduced by Ren and Malik [8], who proposed oversegmentation, i.e., the process of partitioning an image into contiguous regions so that the objects of interest are themselves subdivided into distinct regions, as a preprocessing step to reduce image complexity while preserving most of the structure necessary for segmentation at the scale of interest. Supervoxels group voxels into visually meaningful building blocks of more or less similar size and compactness, depending on the chosen generative approach and parametrization. They reduce data dimensionality with minimal information loss, thereby greatly reducing computational cost of subsequent processing steps, and they allow the computation of local features such as color histograms and texture descriptors. To date, a multitude of different supervoxel algorithms exist which can be categorized by their high-level approach, into e.g., graph-based, density-based and clustering-based algorithms [19].

We use the *SLIC0* variant of the clustering-based algorithm *simple linear iterative clustering* (*SLIC*) [20], due to its comparatively strong performance regarding Boundary Recall and Undersegmentation Error [19], and its memory and runtime efficiency [20]. Its iterative nature enables straightforward runtime restriction and provides direct control over the number of generated superpixels. *SLIC* is an adaptation of k-means clustering with two main distinctions: i) The search space is reduced to a region proportional to the superpixel size, yielding a complexity linear in the number of pixels and independent of superpixel count. ii) The weighted distance measure used combines color and spatial proximity, providing control over size and compactness of the resulting superpixels. The *SLIC0* variant adaptively choses the superpixel compactness, thereby reducing the free parameters to be adjusted to the number of superpixels, or the superpixel size, respectively. The *SLIC0* algorithm was implemented as an ITK (Insight Toolkit) filter and extended to work in three dimensions. Examples of superpixel oversegmentations can be seen in *Fig.2*.

In brightfield and confocal microscopy, tissues or physiological entities are stained; accordingly, color, color gradients or boundaries, and local color patterns are the primary source of information for image interpretation. Therefore, the implemented supervoxel features comprise local and neighborhood color histograms, edge-ness over several spatial scales, as well as texture descriptors (detailed description in *SI Appendix*), yielding a feature vector with a maximum of 138 entries. Feature categories can be disabled by the user to speed up processing.

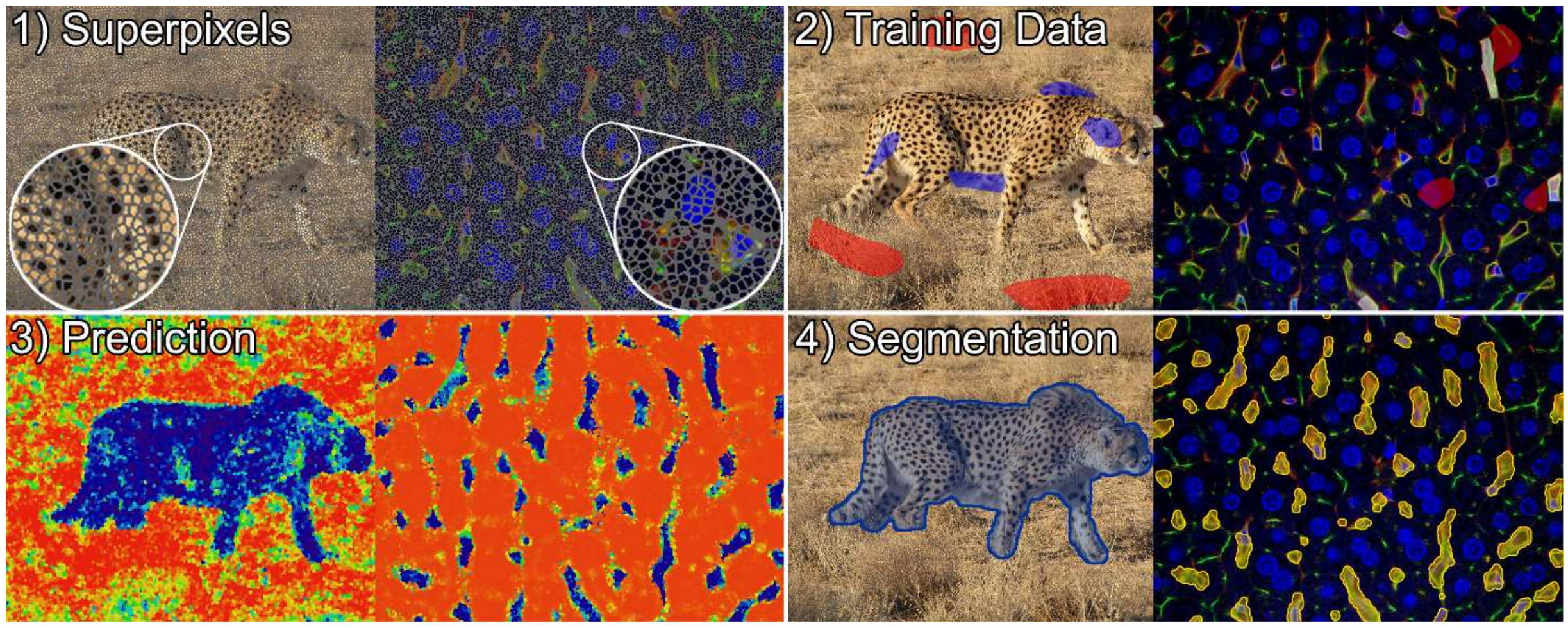

**Fig.2:** Illustration of supervoxel-based image segmentation procedure on an image of a cheetah (left), and a 3D confocal micrograph of mouse liver tissue in which blood vessels were segmented (right). The capillaries in this imaging setup appear yellow-green and usually comprise an unstained lumen as well as few small, elongated nuclei colored blue, which typically belong to endothelial (sinusoidal) cells forming the capillary wall, immune cells residing in the lumen or other non-parenchymal cells. **(1)** Supervoxel outlines are shown in gray. **(2)** Training data for the background class is colored red in both instances, while the foreground class is colored blue in the left and white in the right example. Annotations were selected to represent the characteristic visual features of both classes. For example, annotations of capillaries comprise yellow-colored walls, lumen and enclosed cell nuclei, whereas the background annotations encompass unstained cytoplasm and nuclei of the parenchymal liver cells (blue), bile canaliculi (green) as well as regions close to capillaries to promote learning of exact boundaries. **(3)** Class membership probabilities in the prediction images are illustrated using a color mapping ranging from red (low probability of being foreground) over yellow to blue (high probability of being foreground). **(4)** The segmentation is visualized by a blue overlay on the left, and a yellow overlay on the right.

## Interactive training data generation

For training data generation, TiQuant provides a graphical user interface that allows visualization of and interaction with images. The user can draw directly on the image in order to denote exemplary regions for the classes to be segmented. Supervoxels in the annotated regions are collected and their feature vector together with the annotated class are written into a training database. For examples of how training data annotation is presented in the software, see *Fig.2*.

## Supervoxel-Classification

Random Forests (RF) and Support Vector Machines (SVM) are supported as classifiers to learn the class membership of supervoxels given their feature vectors. We included both classifiers to choose from as they were found to be the two best-performing classifier families when evaluated on a diverse set of learning problems [21]. Furthermore, both classifier families are able to handle non-linear data and have few parameters, however, RFs are faster in regard to training and execution and are therefore set as default.

The user-supplied training data, which is summarized in the training database, is split into a training set and a test set in a 70:30 ratio to allow evaluation of classifier performance (see *Fig.3*). Since training data can be relatively sparse, we use this skewed ratio to ensure a sufficiently large database for training, potentially at the expense of accuracy in assessing classifier performance. Furthermore, the training data is expected to be imbalanced, i.e., that classes are unevenly distributed. To account for this disparity, the initial training-test split, as well as splits performed during cross-validation, are done in a stratified fashion to preserve relative class frequencies. Moreover, in the training phase, the training samples are weighted, where the weight is inversely proportional to the class frequency, and appropriate scoring functions that are insensitive to imbalanced training data are used for classifier evaluation. The feature vectors computed during supervoxel initialization, described in detail in the *SI*, are normalized by subtracting the mean and scaling to unit variance independently for each feature component to ensure comparable feature scales.

Prior to classifier training, hyperparameter optimization of the chosen RF or SVM classifier can optionally be performed to tune parameters that are not directly learned to the learning problem. In order to limit execution time while retaining the explorative quality of an exhaustive Grid Search, the optimization is done using Random Search, which tests a fixed number of parameter settings sampled from given distributions [22]. The search is performed on the training split with 5-fold cross-validation employing the balanced accuracy score [23] to evaluate classifier performance.

As an intermediate step, the (optionally) optimized classifier is trained on the training split and evaluated on the withheld test split to assess its performance. This allows the user to evaluate the suitability of the selected features and classifier, to compare different configurations and to determine whether additional training data is expected to improve performance.

Finally, to make efficient use of the sparse training data, the classifier is trained and calibrated on the entire user-supplied data corpus using the previously determined optimized parametrization. The resulting trained classifier is then used to predict the class membership probability estimates of all supervoxels of the image. Exemplary probability maps are shown in *Fig.2*. Classifier calibration is performed after training using 5-fold cross-validation to provide calibrated probabilistic estimates of class membership as output. In case of SVMs, Platt Scaling [24] is used for this purpose, fitting an additional sigmoid function to map SVM scores to probabilities. RFs, on the other hand, provide probabilistic estimates per default, which can be optionally calibrated using Platt Scaling or the non-parametric Isotonic Regression approach [25]. Empirical results show that SVMs and RFs are among the models that predict the best probabilities after calibration [26]. The Brier score is used to evaluate classifier performance during calibration, since it is a *proper scoring rule* that measures the accuracy of probabilistic predictions [27].

## Segmentation

The probabilistic estimates are mapped to a binary class-membership by applying a threshold, which defaults to the value of 50%. Subsequently, the resulting segmentation can be post-processed using three optional filters. The first post-processing step allows for removal of isolated foreground objects smaller than a specified minimal foreground object size, as well as filling of holes in foreground objects smaller than a minimal hole size. Additionally, a morphological closing operator followed by an opening operator can be applied to smooth surfaces. Specialized label set operators are used that work on the label image representation in which pixel values represent connected objects. These operators preserve the initial label assignment by not merging disconnected foreground objects [28]. Finally, the watershed algorithm may be applied in order to split wrongly connected objects such as nuclei [29]. Two exemplary segmentations are shown in *Fig.2*.

## Color-based image clustering

A common but severe problem hampering efficient batch processing of a large number of images is visual heterogeneity of images due to varying imaging settings and conditions, or systematic visual changes during the imaged process. One approach to solve this problem is image normalization that aims to visually equalize images prior to segmentation and analysis.

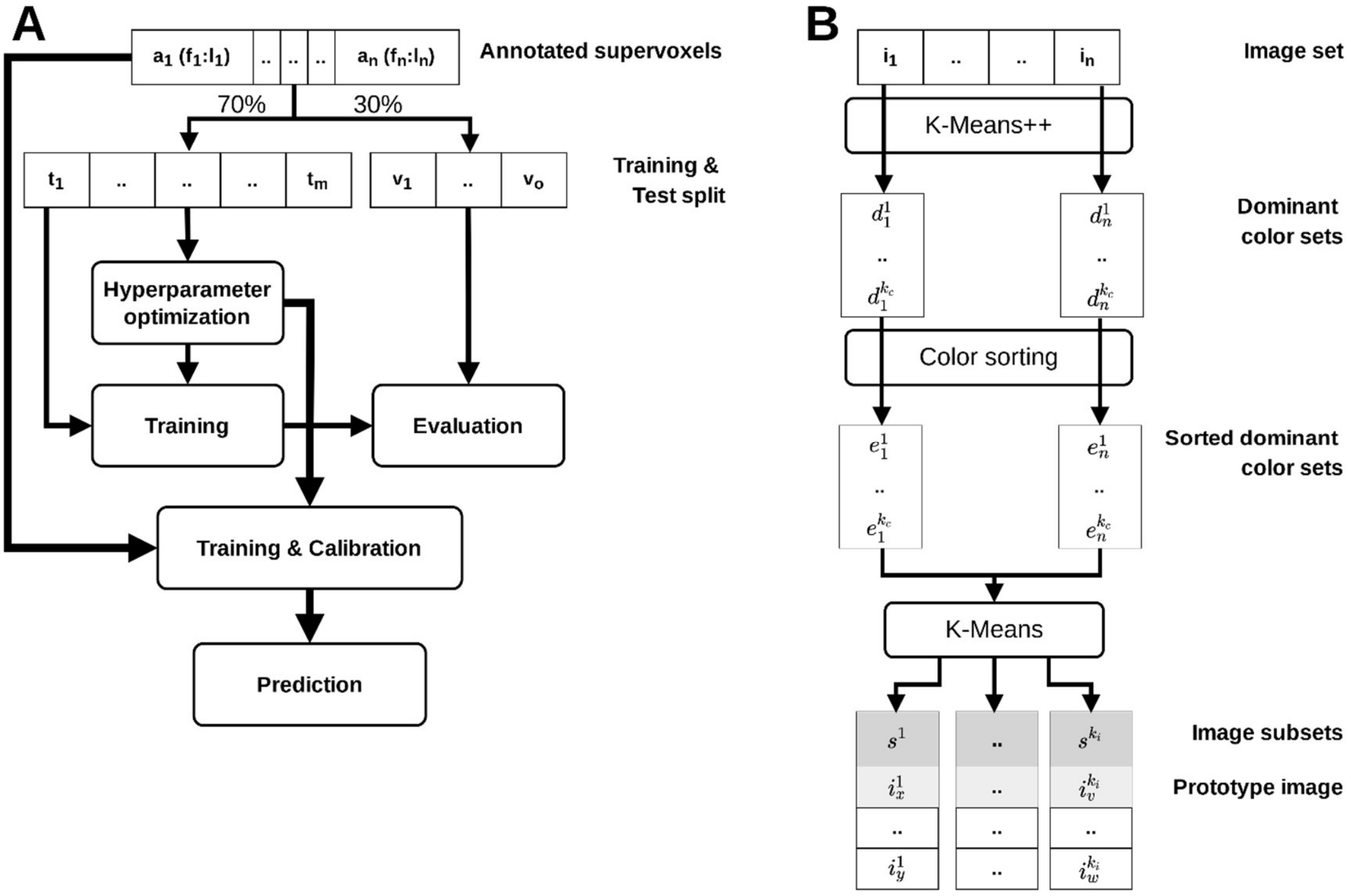

**Fig.3: (A)** Flowchart of the classifier training procedure. The set of annotated supervoxels is split in a 70:30 ratio into a training and test split. The RF or SVM classifier is trained on the training split after optional hyperparameter optimization and evaluated on the test split to allow the user to assess and compare classifier performance. The classifier employed for final prediction uses the optimized parametrization and is trained and calibrated on the entire original set of annotated superpixels. Fig 1A illustrates the integration of the training procedure into the interactive segmentation workflow. **(B)** Flowchart of the color-based image clustering algorithm. For each image, a characteristic set of the $k_c$ most dominant RGB colors is determined by k-means clustering of the RGB pixel values using the k-means++ initialization scheme. To ensure comparability of these representative sets, their RGB entries are sorted by component-wise comparison of R/G/B values. Finally, k-means clustering is applied in the $3 * k_c$-dimensional space of sorted dominant color sets to group the corresponding $n$ images into $k_i$ subsets. For each of these subsets, a prototype image is identified, which is recommended for interactive segmentation. Fig 1B illustrates the integration of the color-based image clustering algorithm into the workflow of guided reuse of interactively trained classifiers.

There is a large number of normalization methods available today, typically aiming at normalizing the intensity distribution of images and/or improving the signal-to-noise ratio, such as variations of adaptive histogram equalization [30], color deconvolution [31], or more recent deep learning based methods [32]. However, the selection of a suitable method as well as its parameterization is non-trivial. Moreover, normalization methods can easily introduce systematic errors in subsequent image analysis if differences in visual appearance originally result from changes in tissue conditions to be analyzed. In this case, normalization could remove properties that actually should be quantified.

We propose a different method to overcome the problem of visual heterogeneity in microscopic images by automatically and robustly grouping images of a given dataset into visually homogeneous subsets. Thereby, we can not only guide the selection of training images, but also facilitate classifier reuse leading to minimized manual annotation effort. The underlying assumption is that minor procedural discrepancies in the image acquisition process introduce systematic changes of coloring. Possible causes comprise slight variations between sample preparation sessions (e.g., affecting staining penetration depth, thus color saturation), minor deviations in imaging settings between imaging sessions (e.g., affecting brightness and contrast), as well as differing conditions of the imaged subjects (e.g., healthy vs. impaired tissue).

In a first step images are analyzed for their dominant colors (see *Fig.3*). This process, also known as palette design, is one of two phases of color quantization, an operation used for e.g., image compression. It has been shown that k-means clustering is an effective method for this task [33, 34]. In order to identify a set of the $k_c$ most dominant colors of an image in RGB color space, each pixel's RGB color vector is interpreted as a data point in 3D space. Instead of starting with fully random cluster centers the k-means++ initialization scheme [35] is used, as it has been demonstrated to improve effectiveness for this task [34]. Elkan's algorithm [36], which is a faster variant of Lloyd's algorithm [37], using the triangular inequality to avoid many distance calculations, is employed to cluster the RGB color data points.

Next, the entries of the dominant color sets are sorted, since their initial order is based on color prevalence, i.e., the number of pixels assigned to a respective color cluster. This ordering, however, is susceptible to changes in image composition, so that, e.g., the size or number of physiological entities influences the rank of the color cluster(s) they are assigned to. Sorting, which is done component-wise on the RGB vector entries, allows comparing the dominant color sets of different images.

Finally, the image set is partitioned into subsets with similar dominant color sets. Our approach extends previous work, in which color moments [38] or histograms [39] were used as image descriptors for k-means based image clustering. Each image is represented by its $3*k_c$ dimensional sorted dominant color set. K-means clustering is applied to this set of data points yielding $k_i$ clusters, minimizing the within-cluster variances of the sorted dominant color sets. The images with the smallest Euclidean distance to their respective cluster center are recommended training images for their image cluster.

# Experiments

## Validation

We validated our interactive segmentation and image set clustering method exemplarily on a set of 22 brightfield micrographs of mouse liver paraffin slices. These images were hand-labeled by a trained expert to automatically extract architectural and process parameters, which were used to parameterize a spatio-temporal tissue model of tissue regeneration after $CCl_4$ intoxication [3]. Tissue slices were immunostained for BrdU positive nuclei to visualize proliferation and typically show a single liver lobule centered by a central vein. Image data was obtained for a control (t=0) and at 7 time points after administration of $CCl_4$, which causes necrotic tissue damage in the area around the central vein. Over the observed time period, this damage regenerates and the lost cells are replenished by cell proliferation of the surviving liver cells.

We reuse this image set to validate our method as it is representative for many image segmentation tasks in a biomedical context, which are characterized by a small total number of available images and heterogeneous image appearance caused by variations in sample preparation, imaging settings, or tissue conditions, or by various other possible technical problems like entrapped air, blurring, or imperfect staining application.

A re-examination of the original nuclei annotations revealed several inaccuracies and inconsistencies (*Fig.4A*). Therefore, two trained experts manually revised all 22 images thoroughly to define a gold standard nuclei annotation. The annotations represent each nucleus as a 2D pixel coordinate. As a measure for segmentation accuracy, we use the $F_1$ score, thereby considering both precision and recall. The number of true positives (tp), false positives (fp) and false negatives (fn) underlying the score are quantified based on an object-wise mapping of the gold standard to the respective segmentation. Specifically, a segmented nucleus is considered a tp if a gold standard nucleus (i.e., its 2D coordinate) lies within the segmented region. Accordingly, a segmented nucleus within which there is no gold standard nucleus is counted as a fp. A fn is registered if there is no segmented nucleus in the immediate neighbourhood (kernel of size 1) of a gold standard nucleus. Based on the $F_1$ scores, we validate and compare the following methods: *i)* original manual annotation (*Fig.4A*), *ii)* our superpixel-based interactive processing approach (*Fig.4B*), *iii)* intra- and *iv)* inter-cluster reuse of interactively pre-trained classifiers as well as *v)* the state-of-the-art tool ilastik [18] as a reference point.

For validation of the superpixel-based approach, each image was partitioned into approximately 50k superpixels of target size 64 pixels (initially 8x8 pixels), and each superpixel was analyzed for all available features, comprising gradient magnitude, Laplacian of Gaussian, local and neighborhood color histograms as well as texture features.

In order to evaluate the suitability of our tool for interactive segmentation, appropriate training data was produced for each of the 22 images by annotation (following workflow *Fig.1A*). Per image, a RF classifier was trained and calibrated after hyperparameter optimization was performed. The trained classifiers were applied to the respective image to predict class membership of all contained superpixels. The resulting segmentations were post-processed by removing objects smaller than three superpixels, smoothing of boundaries of segmented nuclei, and finally by applying the watershed algorithm to split up agglomerations of nuclei into individual objects.

Subsequently, to evaluate how well the trained classifiers generalize to unseen images and whether a restriction to images with similar coloration improves performance, we grouped the image set into $k_i$=6 subsets. The number of clusters was determined experimentally using the Elbow method, a heuristic for determining the optimal number of clusters representing the majority of the variability in a dataset. The number of dominant color sets $k_c$=4 was chosen manually to correspond with the different image compartments of relevant size (i.e., background, nuclei, and two tissue compartments for healthy tissue/necrotic lesion, or glutamine synthetase positive/negative zone, respectively). Per image subset the image with the smallest Euclidean distance to the image cluster center was selected as prototype image. The previously trained classifiers of these prototype images were applied (i) *intra-cluster* to all other images in the respective cluster and (ii) *inter-cluster* to all images not belonging to the respective cluster. Segmentation post-processing was done as described before.

We compare our approach with the established image processing software ilastik. For the results to be comparable, we used ilastik's 'Pixel Classification + Object Classification' workflow. All 37 predefined features were selected for training. The training annotations created with our software were imported and an object size filter was used for segmentation post-processing.

| **Method** | **median($F_1$)** | **$\sigma$($F_1$)** |
|---|---|---|
| manual | 0.9021 | 0.027 |
| interactive segmentation | **0.9297** | 0.013 |
| intra-cluster reuse | 0.9018 | 0.044 |
| inter-cluster reuse | 0.6484 | 0.245 |
| Ilastik [18] | 0.9079 | 0.022 |

**Tab.1:** Comparison of different segmentation methods on 22 images. Bold indicates the best performance.

The summarized results of the method validation are shown in *Tab.1,* while a more detailed per-image/cluster analysis is shown in the *SI Appendix Fig.S3*. The best result was achieved with the fully interactive approach where dedicated training data was provided for each image. The original manual annotation produced comparable results to intra-cluster reuse of classifiers trained on prototype images, indicating that the restriction of classifier reuse to similarly colored images is able to produce human-grade results. However, as the increased variance for classifier reuse indicates, segmentation quality is less reliable. Reusing classifiers on images with differing coloration generally degrades segmentation quality greatly, as expected given the importance of color information. Fully interactive segmentation with ilastik, using dedicated training data for each image, performed slightly better than the pre-trained classifiers on intra-cluster images.

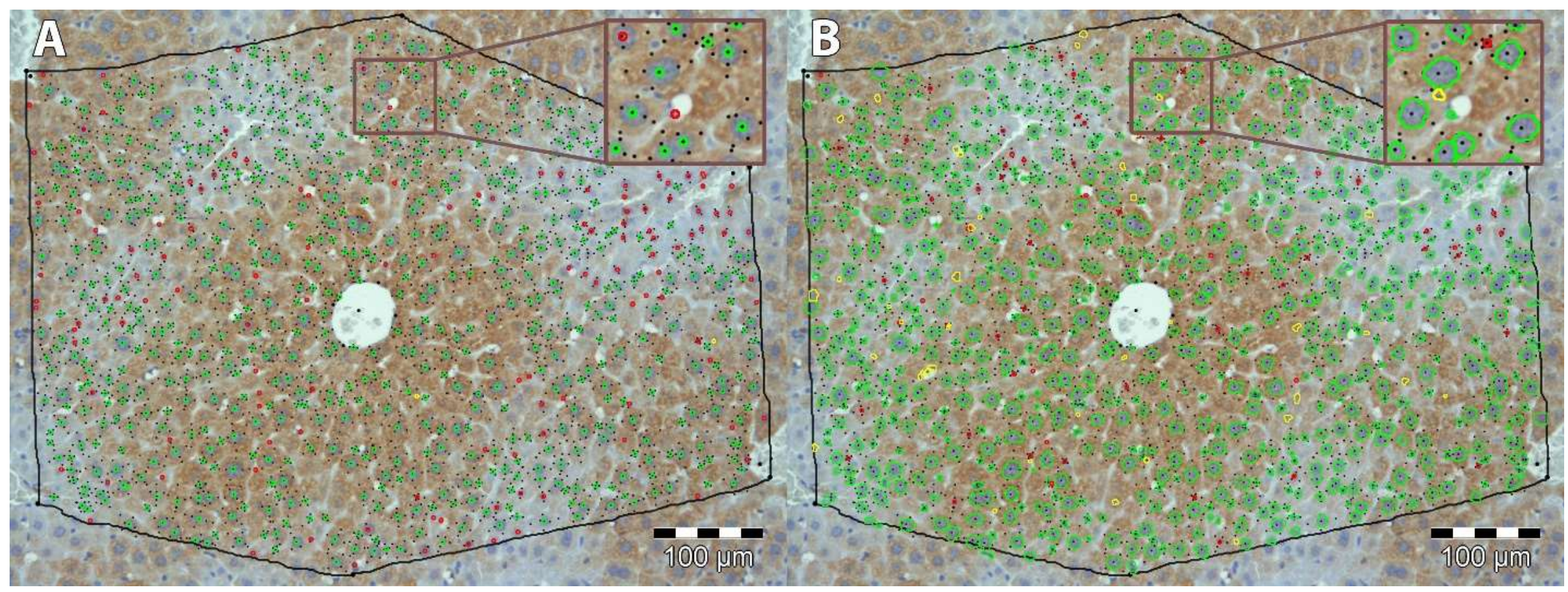

**Fig.4:** Visualization of error classes from comparison with gold standard (GS): **(A)** Comparison of *GS* with the original manual segmentation created by a single human annotator: Correctly segmented nuclei (true positives: bright green); Incorrectly segmented nuclei (false positives: yellow); Nuclei that were falsely not segmented (false negatives: red). **(B)** Comparison of *GS* with a segmentation generated by our interactive segmentation tool: Coloring as in (A). Nuclei outlines in (A) were generated from 2D pixel coordinates provided by the annotator, while outlines in (B) were generated directly from the nuclei segmentation.

### Runtime evaluation

A detailed discussion of aspects of computational performance is given in the *SI Appendix*. The interactive processing steps training and prediction are independent of pixel count and linear in the number of supervoxels. For the segmentation step, runtime scales linear with image size and number of supervoxels. Exemplarily, execution times for an image with 100M pixels and 100k supervoxels are ~2min and 9min for supervoxel generation without and with feature generation. Training database generation, prediction and segmentation were completed in 5sec, 12sec and 51sec, respectively. The analysis was done on an Intel Core i9-7900X with 10 logical cores at a 3.3 GHz clock speed with 64 GB RAM.

## Discussion

Currently, interactive image segmentation often is the optimal image processing approach for applications where a detailed analysis of objects of interest is the aim, but only few images are available, and/or training annotations are not provided. However, these seemingly specific conditions are fulfilled in many cases of biological and medical image processing where often pixel-accurate segmentations of physiological entities are required based on two- or three-dimensional images. This includes the counting and measuring of (sub-)cellular structures and vascular networks in tissue. Custom-made image processing solutions that target specific physiological entities imaged utilizing a specific technical setup are still a very common and widespread approach. Unfortunately, these solutions usually cause considerable development overhead compared to interactive solutions based on machine learning that can be easily adapted to new data through training. Moreover, custom-made solutions require profound expertise in image analysis techniques including knowledge of particular algorithms and the impact of parameter choices. In contrast, interactive image segmentation using machine learning from manual image annotations has a much flatter learning curve and greatly simplifies the process of quantifying the content of complex images enabling users to directly apply their expert knowledge of, for example, biological or medical entities in the analyzed images. The method described in this paper reduces the technical parameters virtually to zero. Only the size of the superpixels has to be set, however, it can be inferred from the size of the structures to be quantified.

Furthermore, interactive segmentation is also applicable for larger datasets consisting of many images, when no ground truth or appropriate pre-trained deep learning models are available. In this case, interactive image segmentation can be useful in producing training data for deep learning with significantly reduced manual effort and thus cost.

The presented supervoxel approach using the *SLIC0* algorithm is especially suited as a basis for responsive, interactive segmentation not only because it is easy to use, but also technically due to its dimensionality reduction characteristic, minimizing processing times and memory consumption. The latter effect is particularly pronounced if the objects of interest are highly resolved. On the other hand, the advantage diminishes if object size approaches those of single pixels. For objects with fluent boundaries or low-resolution images, the alignment of supervoxels to object boundaries can be suboptimal and traditional pixel-level segmentation methods might provide higher segmentation accuracy especially in boundary areas.

The features implemented in the presented work are designed for application to color images and might be less effective in representing learning problems formulated based on grayscale or binary images, for example electron microscopy or x-ray, for which optimized features could be integrated.

Learning from annotations on single images and reusing these interactively trained classifiers for unseen images would further decrease manual effort. However, the downside of learning from such sparse annotations is a decreased generalizability which is especially deteriorating results in case of high color variations between the images to be quantified. We have demonstrated that limiting classifier reuse to similarly colored images significantly increases the average performance on a validation dataset, and thus improves the suitability of interactive image segmentation for processing moderately sized image sets.

## Conclusion

We present a novel approach that combines machine learning based interactive image segmentation using supervoxels with a clustering method for the automated identification of similarly colored images in large image sets which enables a guided reuse of interactively trained classifiers. Our approach solves the problem of deteriorated segmentation and quantification accuracy when reusing trained classifiers which is due to significant color variability prevalent and often unavoidable in biological and medical images. We demonstrate that our interactive image segmentation approach achieves results superior to both manual annotations done by a human expert and also an exemplary popular interactive segmentation tool when dedicated training data is provided for each image (see Tab.1). We show that by limiting classifier reuse to automatically identified images of similar color properties, segmentation accuracy is significantly improved compared to a traditional, non-discriminative reuse. In addition, our approach greatly reduces the necessary training effort. This increase in efficiency facilitates the quantification of much larger numbers of images thereby improving interactive image analysis for recent technological advances like high throughput microscopic imaging or high-resolution video microscopy [40]. Additionally, the presented method can readily be used for a fast generation of training data for deep learning approaches. In summary, our approach opens up new fields of application for interactive image analysis especially in but not limited to a biological and medical context. The provided free software makes the presented methods easily and readily usable.

Supporting Information (SI) for publication:

# Guided interactive image segmentation using machine learning and color-based image set clustering

## Superpixel features

The superpixel features comprise local and neighborhood color histograms, the averaged gradient magnitude and Laplacian of Gaussian for several sigmas, as well as texture descriptors. Histograms are calculated independently for the red, green and blue channel with each histogram having 16 bins, thus representing 4096 colors. For local histograms all pixels of a superpixel are taken into account. The neighborhood histograms additionally account for all pixels of all neighboring superpixels, thereby providing context. As a means for isotropic edge detection the gradient magnitude is computed separately for the red, green and blue channel after applying gaussian smoothing with variable sigma (0.5, 1, 2, 4, 7, 10). By varying sigma, edges of different scales are accentuated. The gradient magnitude is averaged over all pixels of a superpixel per color channel and sigma. As a second edge detection method based on the second derivative the Laplacian of Gaussian (LoG) is computed for each of the color channels and with variable sigma. The LoG is averaged over all pixels of a superpixel per color channel and sigma. Additionally, a set of six texture describing features is computed. Therefore, the RGB image is converted to a grey-level image which corresponds to the luma component Y of the YIQ color space. Based on this a grey-level co-occurrence matrix is computed which is then used to calculate the texture features [1]. Subsequently a texture feature subset comprising Inertia, Cluster Shade, Cluster Prominence, Inverse Difference Moment, Energy and Entropy as proposed by [2] is computed.

In summary, the mentioned features constitute a feature vector of 48 entries per local and neighborhood color histograms each, 18 entries per gradient magnitude and LoG values each as well as six texture descriptor entries, yielding a feature vector with a total of 138 entries to characterize each superpixel.

## Performance evaluation

We analyzed the influence of image size and superpixel number on execution times. In the first scenario the image size is successively upscaled from a 1250 x 1250 pixels image to a 15k x 15k pixel image yielding a 144-fold increase in pixel number while working with a constant number of 100k superpixels. The second scenario image size remains constant with 5000 x 5000 pixels but the number of superpixels increases stepwise from 110,835, resulting from superpixels with a target size of 15 x 15 pixels, to 4,471,427, resulting from superpixels of size 2 x 2 pixels, yielding an increase by a factor of 40.3. We evaluated for each scenario runtime of the interactive steps training, prediction and segmentation. Training used in all instances the same (scaled) training masks. Random Forest without optimization and calibration was used as classifier. The analysis was done on a Intel Core i9-7900X with 10 logical cores at a 3.3 GHz clock speed with 64 GB RAM.

Trivially, the complexity of the training step which maps the user provided training masks to superpixels and compiles the training database from those is independent from the number of pixels. The linear increase by a factor of approx. 2.6 from smallest to largest image shown in Fig.S2 can be attributed to linearly increasing image loading times. Similarly, prediction, which encompasses learning and application of the classifier to each superpixel of the image, is independent of image size as it operates directly on superpixel level. The fluctuations seen in Fig.S1 between 4.8 sec and 11.41 sec are due to differences in classifier parameterization arising during learning. Segmentation is of linear complexity regarding image size, which is confirmed showing an increase by a factor of 27.9 from 3.8 sec for the smallest to 106.1 sec for the largest image. Complexity of training, prediction and segmentation is linear in the number of superpixels due to sequential processing, which is confirmed in Fig.S2. Training times increase 28.1-fold from 3.9 sec to 109.8 sec, prediction times increase 23.5-fold from 7.9 sec to 185.6 sec and segmentation processing times increase 9.8-fold from 17.6 sec to 171.6 sec.

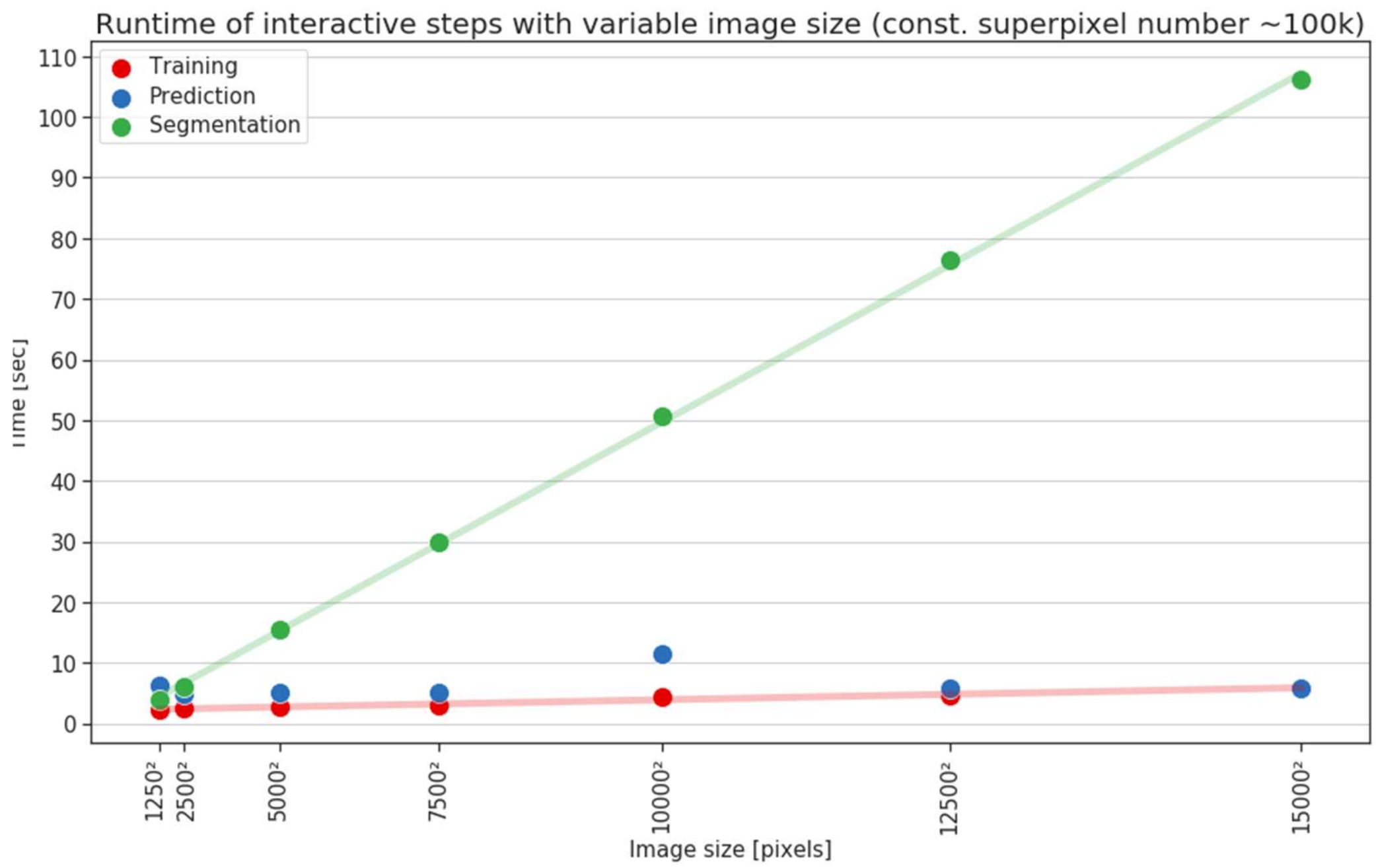

**Fig.S1:** Effect of image size on runtime of interactive processing steps. The training procedure involves identification of annotated superpixels and compilation of the training database. Prediction encompasses fitting of the classifier and prediction of class-membership probabilities for all superpixels of the image. In the segmentation step the final image is produced given the prediction results, no post-processing such as size thresholding, smoothing or watershed is applied. In all instances the same set of training annotations was used, scaled to the respective image size.

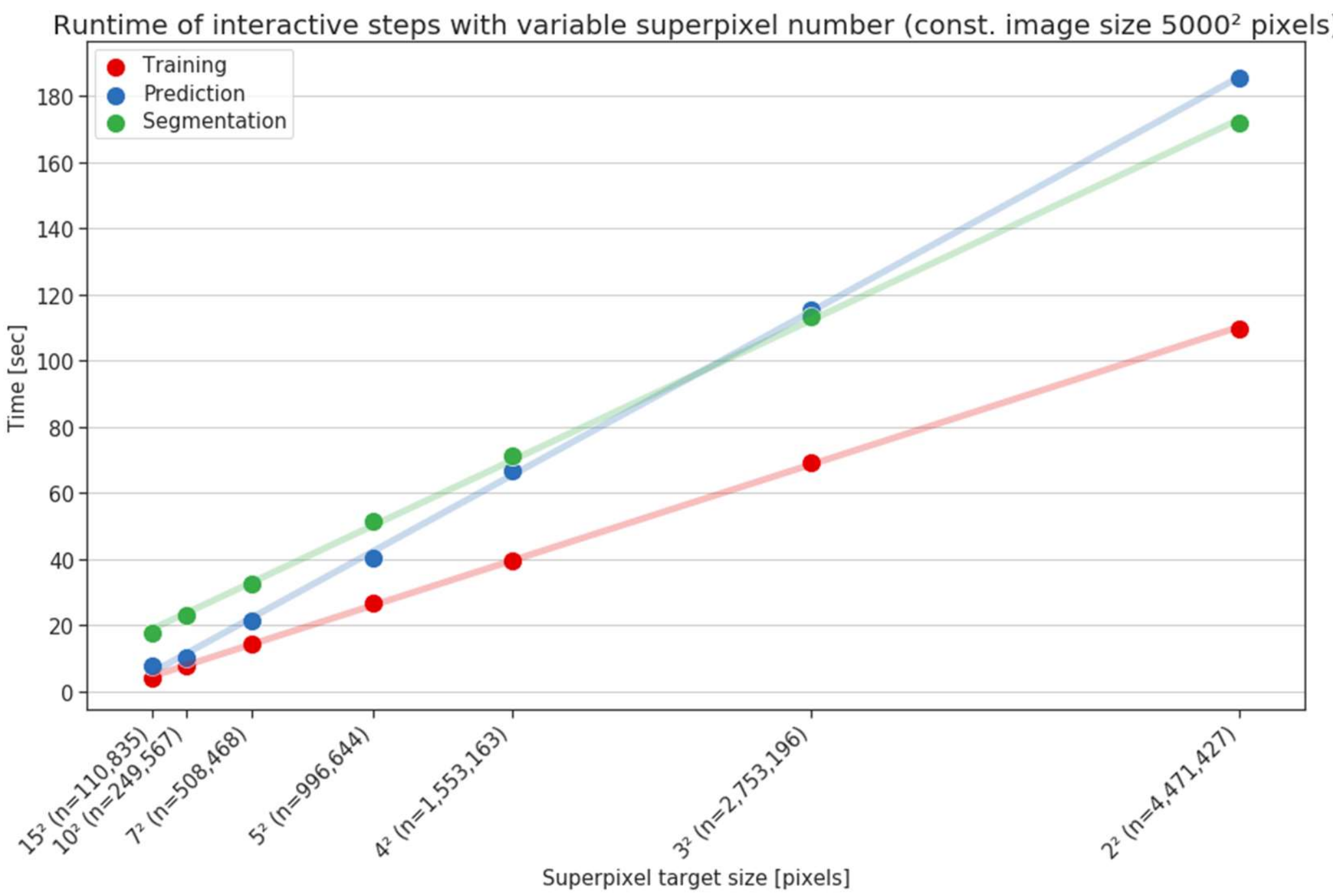

**Fig.S2:** Effect of number of superpixels on runtime of interactive processing steps. In all instances the same set of training annotations was used.

## Comparison of different segmentation methods

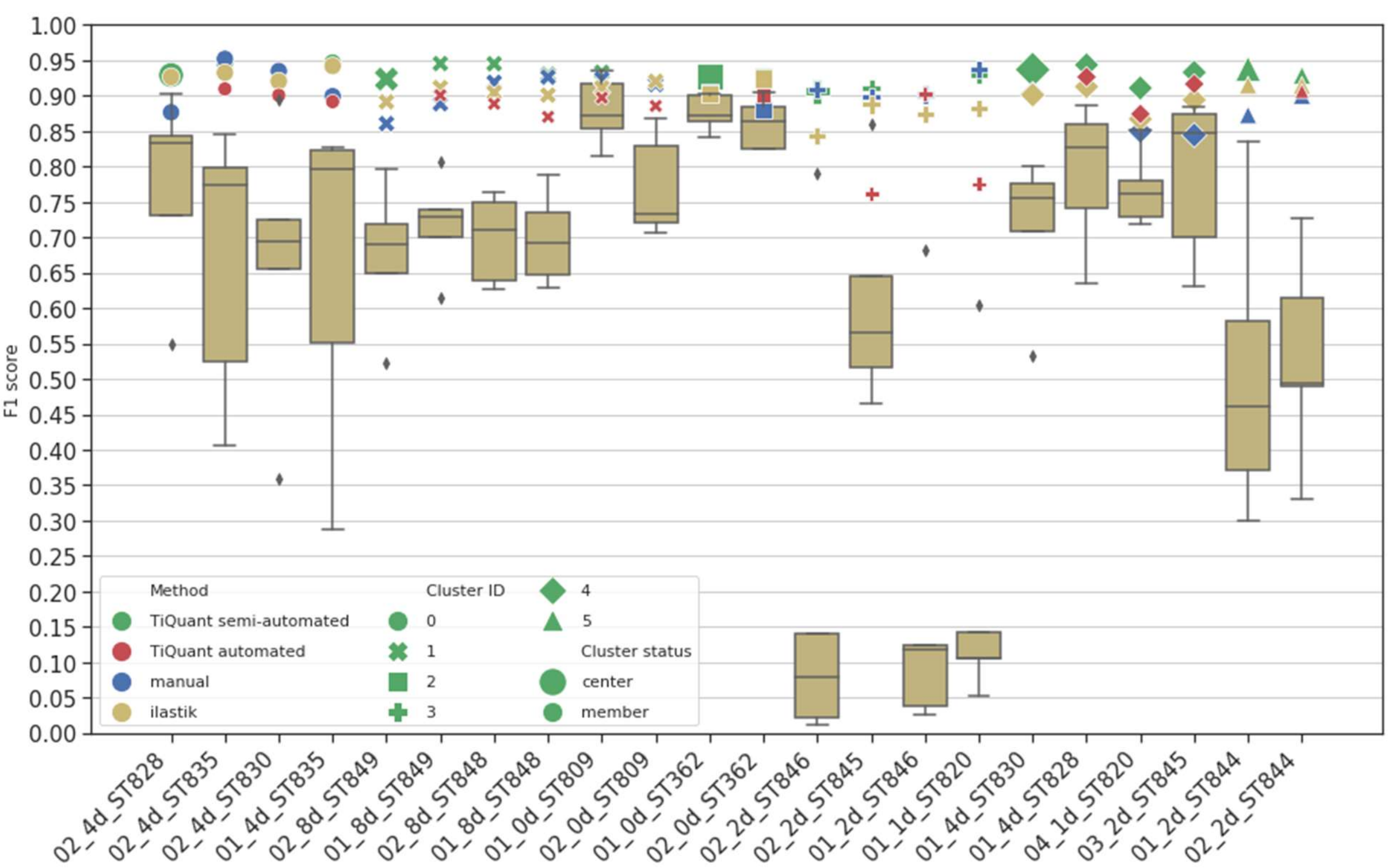

**Fig.S3:** Per image comparison of different segmentation methods: Manual annotation (blue), semi-automated superpixel-based processing (green), intra-cluster reuse of pre-trained classifier (red), inter-cluster reuse of pre-trained classifiers (box) and ilastik's 'Pixel Classification + Object Classification' workflow [3] (yellow). For intra-cluster performance analysis, the classifier trained on the cluster prototype image (leftmost) was reused on the remaining images of the respective cluster. Therefore, the cluster prototype image has no intra-cluster score. Inter-cluster performances are given as a boxplot summarizing the performance of classifiers of all other clusters.

# TiQuant 2.0 User Guide

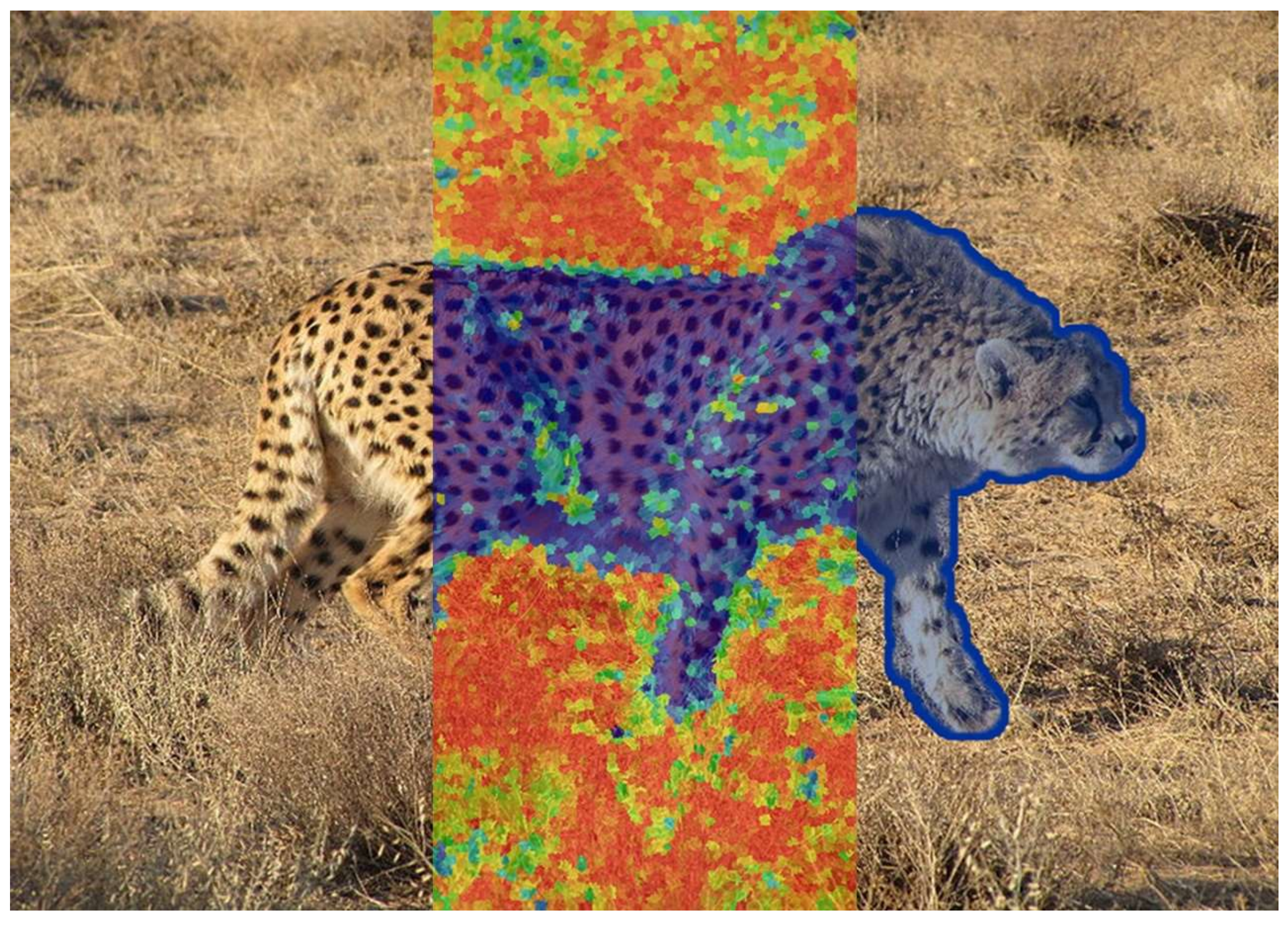

*TiQuant software version: 2.0*

*User Guide version: 2.0*

# Introduction

This document is a user guide / tutorial for the image processing and quantification tool *TiQuant* 2.0. TiQuant can be used as standalone software, but is also integrated with *TiSim* in an image processing and tissue modeling framework developed cooperatively by the groups of Stefan Hoehme at the Institute of Computer Science, University of Leipzig, Germany, and Dirk Drasdo at INRIA, Paris, France.

The user guide includes information on the 'Software design', 'Required hard- and software setup' and provides an 'Installation Guide'. In the chapter 'Working with TiQuant in a nutshell' the Graphical User Interface (GUI) and the batch processing mode are discussed.
The remainder of the user guide is split into three parts:
In the first part usage of a color-based image dataset clustering method is explained that can be used to partition a dataset consisting of images, ideally imaged under comparable conditions using a standard protocol, into subsets of similarly colored images.
The second section covers the versatile superpixel-based image processing tools, which can be used to segment various objects of interest (e.g. nuclei, tumor tissue, blood vessels, ..) in divers types of images (2D whole slide scans, 2D/3D confocal micrographs, satellite images, ..). Besides detailed instructions for the procedural steps a general workflow section characterizes the process of image segmentation.
The third part details all pipeline-based image processing and quantification functionalities that are specifically tailored to 3D confocal micrographs of mouse liver with the staining setup detailed in [1]. Their parameters are explained and it includes comments on how to adjust parameters to improve segmentation quality. It also provides an extensive list of all quantifiable parameters and lastly in the 'Recipe for test dataset' section a concrete step-by-step guide on how to use TiQuant with one of the provided test data sets to produce an exemplary high-quality segmentation.

# Developers and License

**Contributors (Listed alphabetically):**

Dirk Drasdo, Adrian Friebel, Stefan Hoehme, Tim Johann

TiQuant uses external libraries, which are included in the binary package. These libraries have not been changed in source code, but have been built in ways specific to our needs. Below, all external libraries used by TiQuant are listed together with the location of their source code and the corresponding open-source licenses under which they are distributed:

| Library | Version | Homepage | License |
|---|---|---|---|
| Qt | 5.14.1 | http://www.qt-project.org | GNU LGPL v3 |
| ITK | 4.9.1 | http://www.itk.org | Apache License Version 2.0 |
| VTK | 7.1.1 | http://www.vtk.org | BSD 3-clause License |
| Python | 3.6.8 | https://www.python.org/ | GPL-compatible |

# Software design

TiQuant is written in C++ and Python and makes use of the C++ open-source libraries Qt, ITK and VTK as well as the Python packages NumPy, SciPy, matplotlib and scikit-learn. Qt is a widely used framework allowing for cross-platform Graphical User Interface (GUI) development. The Insight Segmentation and Registration Toolkit (ITK) is a well-maintained and documented, cross-platform library that provides an extensive suite of software tools for image analysis. It employs leading-edge algorithms for registration and segmentation of multi-dimensional data. Provided by the same company, Kitware Inc., the Visualization Toolkit (VTK) is a freely available framework for 3D computer graphics, image processing and visualization. A C++ - Python interface enables TiQuant to utilize the well maintained, open-source machine learning package scikit-learn, which provides efficient tools for data mining and analysis such as Clustering, Classification and Regression methods. Scikit-learn is built on the open-source packages NumPy, SciPy and matplotlib, which are also used by TiQuant for scikit-learn related data handling and plotting.

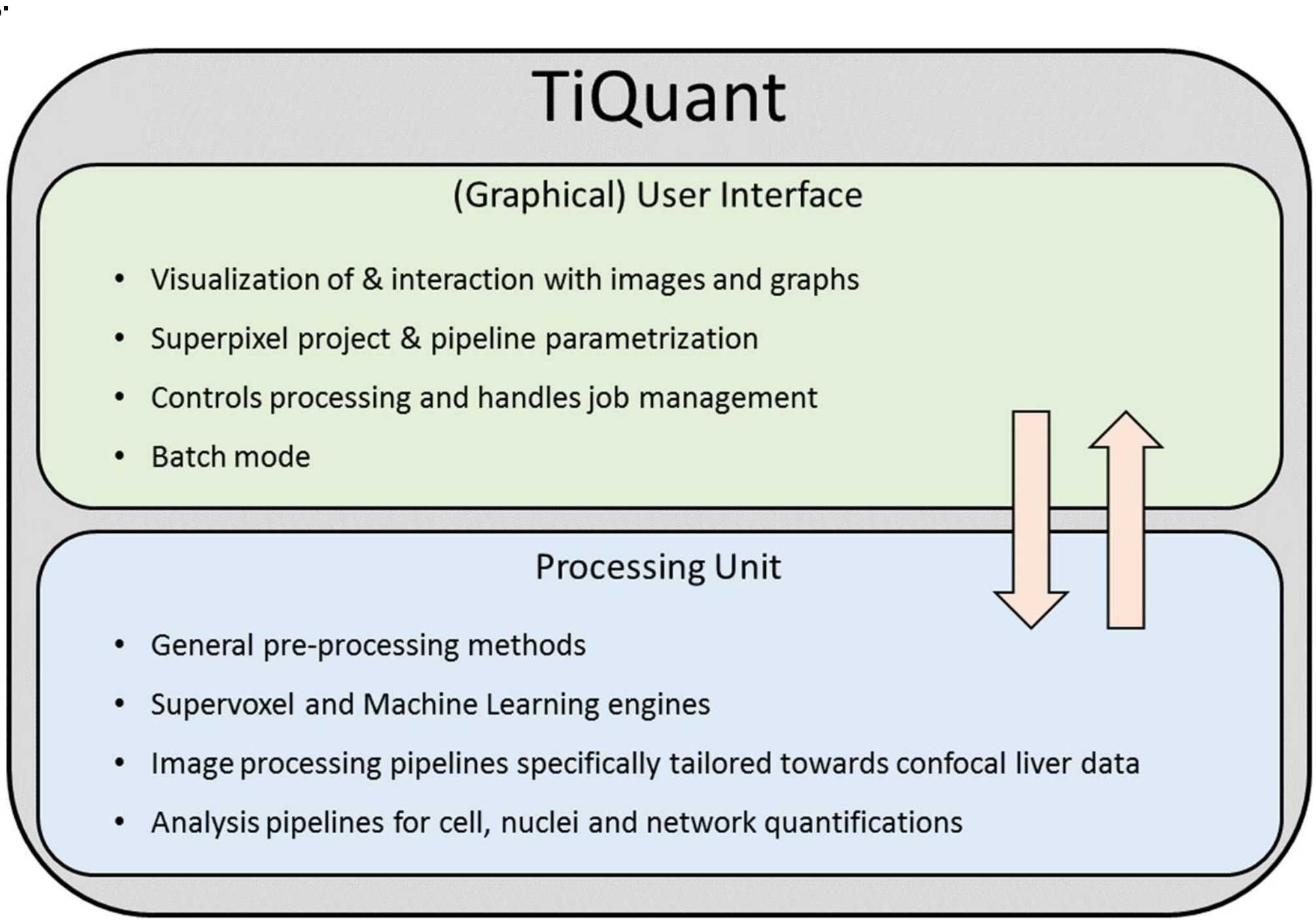

*Illustration of modular design of TiQuant*

TiQuant is of modular design, separating the Graphical User Interface (GUI), which handles all user interactions, from the core Processing Unit (PU). These two building blocks inform each other to e.g. parametrize and execute processing pipelines or show results of processing.

## User Interface

The GUI-frontend uses the Qt and VTK libraries to provide a user interface that enables specification of parameter values, handles interactions with images, e.g. training data production or seed point selection, and graphs, e.g. data inspection. Furthermore, it allows control over pipeline execution, job queuing and batch mode processing. Batch jobs can be started using a command line interface.

## Processing Unit

The PU encompasses, as the name suggests, all methods that process data. These methods can be grouped into one of five categories: **1)** Color-based clustering which is a dataset-level pre-processing procedure partitioning the images of a dataset into subsets of similarly colored images. **2)** Pre- processing procedures on the image-level, which usually perform a single, concrete, optional pre-processing step, such as cropping or contrast limited histogram equalization (CLAHE). **3)** The supervoxel framework suited to segment objects of interest (e.g. nuclei, tumor tissue, blood vessels, ..) in various types of images (2D whole slide scans, 2D/3D confocal micrographs, satellite images, ..) by decomposing the image into perceptually homogeneous regions (superpixels) and subsequent learning of superpixel class-memberships using machine learning methods. **4)** Several processing pipelines for segmentation and reconstruction of veins, necrotic regions, nuclei, bile and sinusoidal networks as well as cells. They are specifically tailored to 3D confocal micrographs of mouse liver with the staining setup detailed in [1] but were successfully applied to rat, pig and human datasets. **5)** Several quantification pipelines that use segmentations obtained by aforementioned pipelines and analyse nuclei, cells, bile and sinusoidal networks for a variety of parameters.

# Required hard- and software setup

1. A computer (64 bit, preferably multi-processor, minimum of 16 GB RAM, 620 MB free disc space) with a Unix/Linux, Microsoft Windows or MacOS operating system.
2. Installation of TiQuant, which is available from our homepage (https://www.hoehme.com/ or http://msysbio.com/). ***For installation instructions, please refer to the next section.***
3. Optionally installation of image processing software capable of reading and showing 3D image stacks (e.g. ImageJ, which can be downloaded here http://rsb.info.nih.gov/ij/download.html).

# Installation Guide

## Windows

1. Download the Windows version (https://www.hoehme.com/software/tiquant).
2. Unzip the downloaded archive.
3. In the unzipped TiQuant folder you will find TiQuant.bat. Double click it to start TiQuant with its graphical user interface enabled. A warning might pop up, click on 'More information' -> 'Start'.

## Linux

1. Download the Linux version (https://www.hoehme.com/software/tiquant).
2. Unzip the downloaded archive.
3. Open the command line, navigate to the unzipped TiQuant folder and use ./TiQuant.sh to start the application. Starting TiQuant.sh using a file browser will work on most desktops.
4. TiQuant is known to run under Fedora Linux 30. It might work on other distributions but is not guaranteed to.

# Working with TiQuant in a nutshell

## Graphical user interface

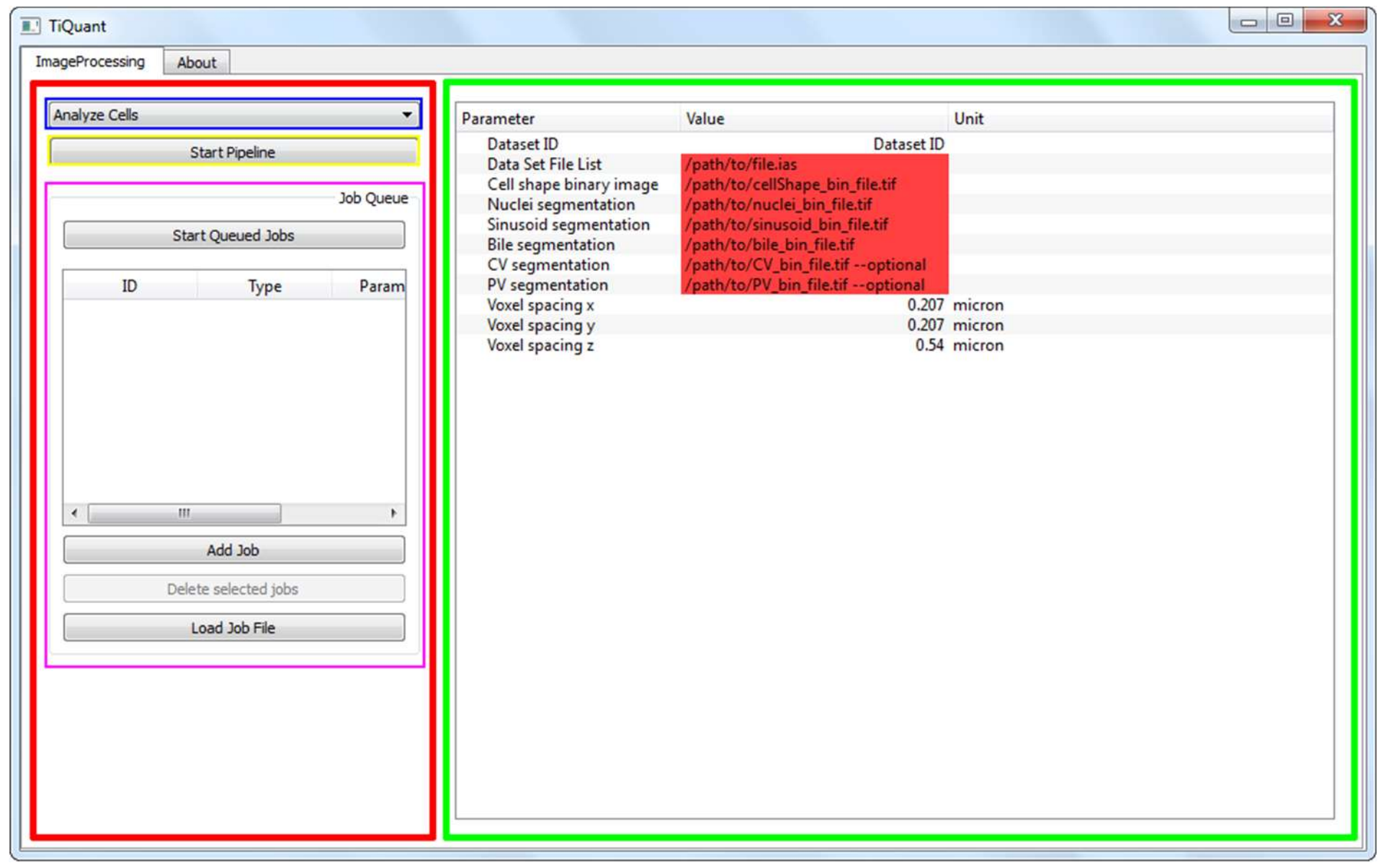

*TiQuant main window*

The user interface of TiQuant is split into two functional panels.

In the left panel (section outlined in red), the processing pipeline selection menu is located (blue), as well as buttons for starting a pipeline (yellow), adding pipeline 'jobs' to a queue, and starting the latter (magenta).

In the right panel (green), a table is shown that allows for parameterization of the currently selected pipeline. Parameter values are editable by double clicking in the corresponding field in the 'Value' column. In case the parameter is a filename, a double click will enable a small button in the value field, which will open a file browser.

After adjustment of all parameters, a pipeline can either be started by clicking 'Start pipeline' or added to the processing queue by clicking 'Add Job' in the left panel. The latter will not start the pipeline immediately, but queue it for later sequential processing. When all jobs have been added, processing of the queue can be started by clicking the 'Start queued jobs' button. Alternatively, a job queue can be loaded by clicking the 'Load Job File' button. The file manager dialog will show TiQuant's job file directory. The default job file is named 'userJobsToDo.txt'.

## Batch processing

A preexisting job queue can be started from command line by executing the following command in the shell:

- **Windows:**
  - Using the convenience script TiQuant_BatchMode.bat, which will spawn a console and start batch job processing, if jobs are saved to default directory (see below)
  - Directly from command line using either the convenience script above or directly ba typing CellSysUI.exe --batch path/to/job_file.txt
  - By default Job files are stored in the user directory of your windows installation, e.g. C:\Users\YourWindowsUserName\AppData\TiQuant\JobManager
- **Linux:**
  - ./TiQuant.sh --batch $HOME/.TiQuant/JobManager/userJobsToDo.txt

# Dataset color clustering

***Required Preparations:***

The algorithm expects the images (2D or 3D) of the dataset to be organized in a common root folder, with each image located in its own sub-folder.

***Procedure:***

1. Select 'ImageProcessing' tab.
2. Select 'Dataset color clustering' in the pipeline selection menu.
3. Set 'I/O -> Dataset root path' to the location of the common root folder in which the subfolders containing the images to partition are stored.
4. Edit 'I/O -> Subfolder name format' to reflect the naming scheme of the subfolders containing the images. The name format has to be given as a regular expression. For explanation and examples see [5].
5. Edit 'I/O -> File name format' to represent file naming scheme. This is given as above using regular expressions.
6. Set 'I/O -> Results path' to a folder in which the results will be stored. This folder needs to be empty, in case it is not, a new results subfolder will be automatically created in the specified directory.
7. Activate 'I/O -> Copy images to result dir', if the images should be copied to the results directory and automatically organized according to their cluster-membership. Please make sure, that you have enough disk space. In any case the resulting image specific color clusters as well as the resulting image clusters are stored in text files and a PDF is created to document the results.
8. Set 'Parameters -> Number palette colors' to the number of colors, each image should be characterized by. This should approximately reflect the distinctively colored compartments in the image (e.g. background, tissue, nuclei, blood vessels, etc).
9. Activate 'Parameters -> Exclude black areas' and 'Parameters -> Exclude white areas' if you there are black/white regions in the image (e.g. background or annotations) that should be excluded when computing color palettes.
10. Adjust 'Parameters -> Resampling factor' to balance accuracy and performance. This factor will be used to downscale the images along each dimension before color palette computation. Higher factors improve performance but might obscure small features (e.g. nuclei).

11. Set 'Parameters -> Number image clusters' to the number of subsets the images should be partitioned in. Since it might be difficult to guess an appropriate value, there is an elbow plot in the final PDF that helps to identify it for a subsequent execution of the pipeline.
12. Once you are finished with configuration of the pipeline click the 'Start Pipeline' button. After execution the results folder is populated with subdirectories representing the image subsets. The subset folders in turn contain subdirectories for each image belonging to the subset. In these image folders the image itself is located, if image copy was activated, as well as a text file holding the extracted color palette for the image with its RGB values as columns and individual colors as rows. Additionally, the clusters.txt file in the root result directory enumerates the image subsets and the PDF file gives a visual summary of the results.

***Estimated execution time:***

On an eight core i7-9700K CPU with 3.60GHz and 16GB RAM for a dataset of 22 images with 2080×1544 resolution and settings set to 4 palette colors, resampling factor of 4 and 6 image clusters computation took ~3 min.

# Superpixel-based image processing

## Workflow

In the initial preparation step an oversegmentation of the image into superpixels is produced using the SLIC0 algorithm [3]. The only adjustable parameter here is the target size of the superpixels, which will determine how well they will be able to represent objects in the image. Afterwards, features based on color, gradients and texture will be computed for each of the superpixels.

The results of the preparation step are encapsulated in a so called superpixel project which is the basis for following training, prediction and segmentation steps. During training the user marks sample regions within the image by a drawing interaction and assigns them to either the foreground (objects the user is interested in) or background (everything else) class. The sample regions should be representative in the sense that they should reflect the visual variability that each class exhibits (e.g. background regions with different noise levels, the different structures that may comprise the background). To finalise the training procedure the user compiles a training database from the training regions, which saves sample regions and database in a so called training project. Subsequently, a classifier is trained on the training database and predicts afterwards for each superpixel a class-membership probability. The result will be saved in a prediction project. In a last step a segmentation can be generated from the class-membership probabilities, which is saved in a segmentation project.

It is recommended to start with a training project with few and small training regions, generate a prediction and optionally a segmentation to get a better idea of its quality. Usually, there are misclassifications and the result can be improved. To this end the user creates a new training project, wherein a few representative samples of the wrongly classified regions are added to the existing training sample regions with there correct class assigned. After creating a thereon based prediction and segmentation the classification of the newly added (and similar) regions should be improved.

By repeating this procedure of evaluating a resulting segmentation, identifying misclassified regions, creating a new training project by adding samples from those regions to the latest training project, and creating a prediction and segmentation based on the amended training data, the classification result can be incrementally improved to a point where the objects of interest are well-classified. Of course, there will be cases where a satisfying result cannot be achieved for various reasons (e.g. image data not well described by available features, data too complex for classifier).

To segment several images of the same type (same microscopy technique, similar magnification & staining, similar visual appearance in general) it is possible to import a trained classifier that lead to a satisfying segmentation on one of the images and use it directly for prediction on the other images. If the images are similar enough this approach eliminates the need to train a classifier for every image. In case the imported classifier doesn't perform well on the new image because of few misclassifications (e.g. due to different background noise levels) it is also possible to import the training database that led to the previously imported classifier and amend the training database by the necessary corrections.

## Required data

The superpixel-based pipelines accept 2D/3D grayscale/RGB images in TIFF format.

## Superpixel Preparation

***Example:***

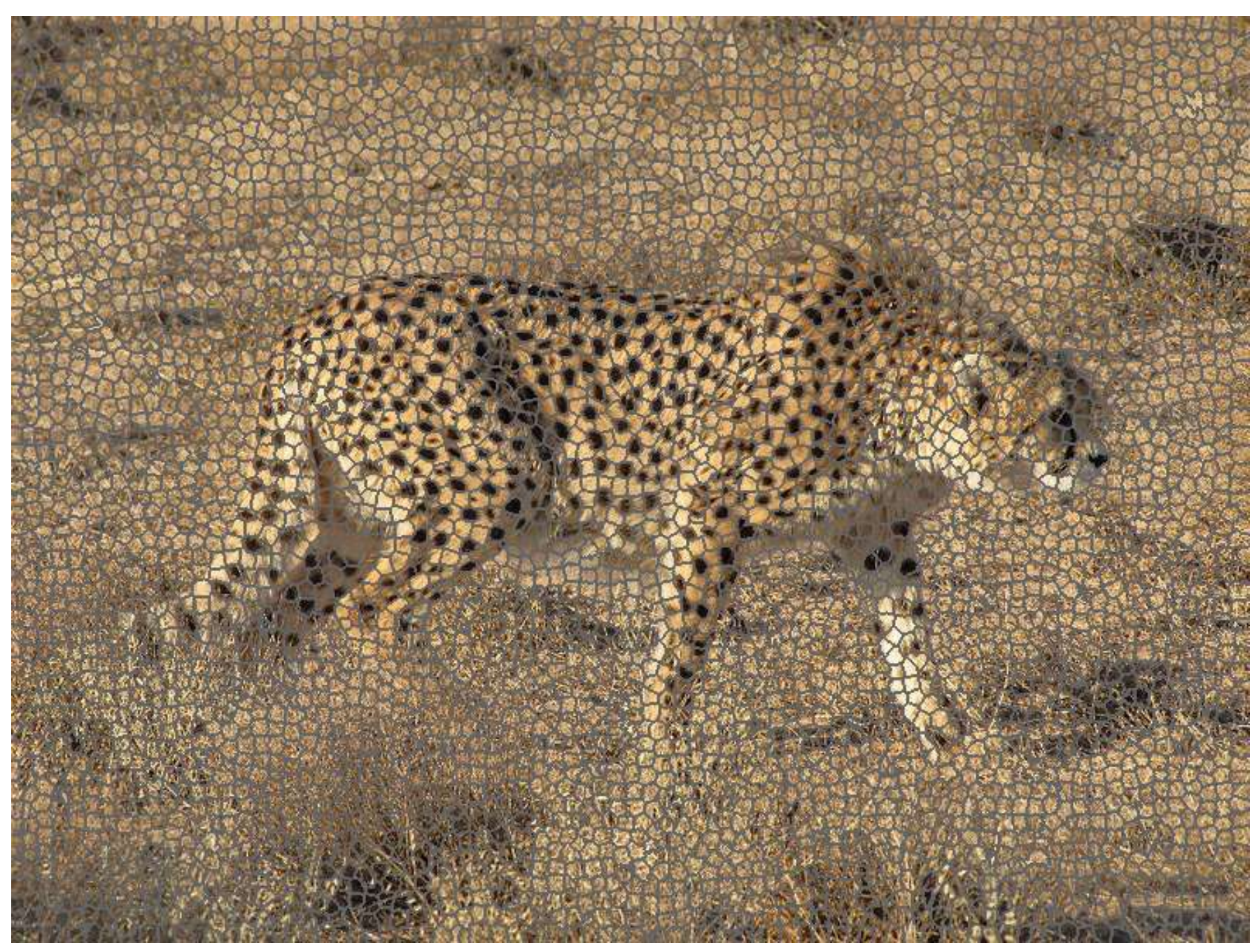

*Original image overlaid with superpixel borders (gray)*

***Required Preparations:***

2D/3D grayscale/RGB images saved in TIFF format.

***Procedure:***

1. Select ‘ImageProcessing’ tab.
2. Select ‘Superpixel Preparation’ in the pipeline selection menu.
3. To create a new superpixel project from scratch select ‘With Superpixel-Partitioning’ under ‘Preparation -> Start configuration’.
4. Set ‘Preparation -> With Superpixel Partitioning -> Reader -> Image filename’ to location of raw image. In case the image is not located in an otherwise empty directory, a dialog will help you to set up a project folder containing only a copy of the image.
5. Adjust ‘Preparation -> With Superpixel Partitioning -> Reader -> Image voxel spacing x/y/z’ for every axis, if necessary. This reflects the physical size of a pixel/voxel and will be taken into account during superpixel size calculation.
6. Set the target superpixel size along the image axes using ‘Preparation -> With Superpixel Partitioning -> Reader -> Superpixel spacing x/y/z’ parameters. Sizes of individual superpixels might vary, as they try to align with image features. As a rule of thumb, superpixel size should be smaller than half the size of the smallest/thinnest part of the objects of interest. This guarantees, that the superpixels fit nicely into the objects of interest, and are able to align well with the color boundaries. However, bigger superpixels might work well too, depending on the tasks, and come with the benefit of shorter processing times. It should be avoided to use sizes of less than five pixels, since features get less meaningful respectively the classification problem becomes harder.

7. Select the features you want to add to the superpixel project under 'Feature Selection -> Unary Features'. It is recommended to use at least the Red/Green/Blue Histograms. Texture features, which describe to some extent spatial arrangement/pattern of colors in the superpixels, are computationally demanding and might take a substantial amount of time to calculate if you are working on a project with many superpixels (>100k). To see the number of superpixels of an already existing superpixel project select 'Superpixel Segmentation' in the pipeline selection menu, open the file browser of 'Superpixel project directory', and navigate with it to the superpixel project in question. On the right side of the file browser general information regarding the project will be summarized, including 'Superpixel Root Project' -> 'Superpixel data' -> 'Superpixel number'.
8. Once you are finished with configuration of the pipeline click the 'Start Pipeline' button. After execution the superpixel project folder is populated with files. For illustrative purpose the file 'superPrep_labelObjectOverlay.tif' is generated showing the raw data overlaid with superpixel boundaries.

***Note:***

Inspection of the resulting 'superPrep_labelObjectOverlay.tif' file located in the project directory using the image viewer of your choice (ImageJ recommended) can help to decide, whether superpixels are of appropriate size, and as that align well with color borders, especially where objects of interest are concerned.

***Estimated execution time:***

On a six core i7-5930k CPU with 3.50GHz and 64GB RAM for an 5000×5000 pixel image and a superpixel size of 10×10 pixel (~250k superpixel) and R/G/B gradient magnitude and R/G/B histogram features activated it took ~23 min.

## Add Features to Superpixel Project

***Required Preparations:***

A superpixel project (created with 'Superpixel Preparation' pipeline).

***Procedure:***

1. Select 'ImageProcessing' tab.

2. Select ‘Superpixel Preparation’ in the pipeline selection menu.
3. In case you want to add features to an already existing superpixel project, select ‘Add features only’ under ‘Preparation -> Start configuration’.
4. Specify under ‘Without Superpixel Partitioning -> Superpixel Project Reader -> Superpixel project directory’ the superpixel project you want to add features to.
5. Select under ‘Feature Selection -> Unary Features’ the features you want to add to the superpixel project. Features already present in the project will be displayed as grayed-out and selected.
6. Once you are finished with configuration of the pipeline click the ‘Start Pipeline’ button. The superpixel project will be expanded by the features to be added.

***Estimated execution time:***

On a six core i7-5930k CPU with 3.50GHz and 64GB RAM for an 5000×5000 pixel image and a superpixel size of 10×10 pixel (~250k superpixel) adding R/G/B gradient magnitude and R/G/B histogram features it took ~12 min.

## Create Training Data

***Example:***

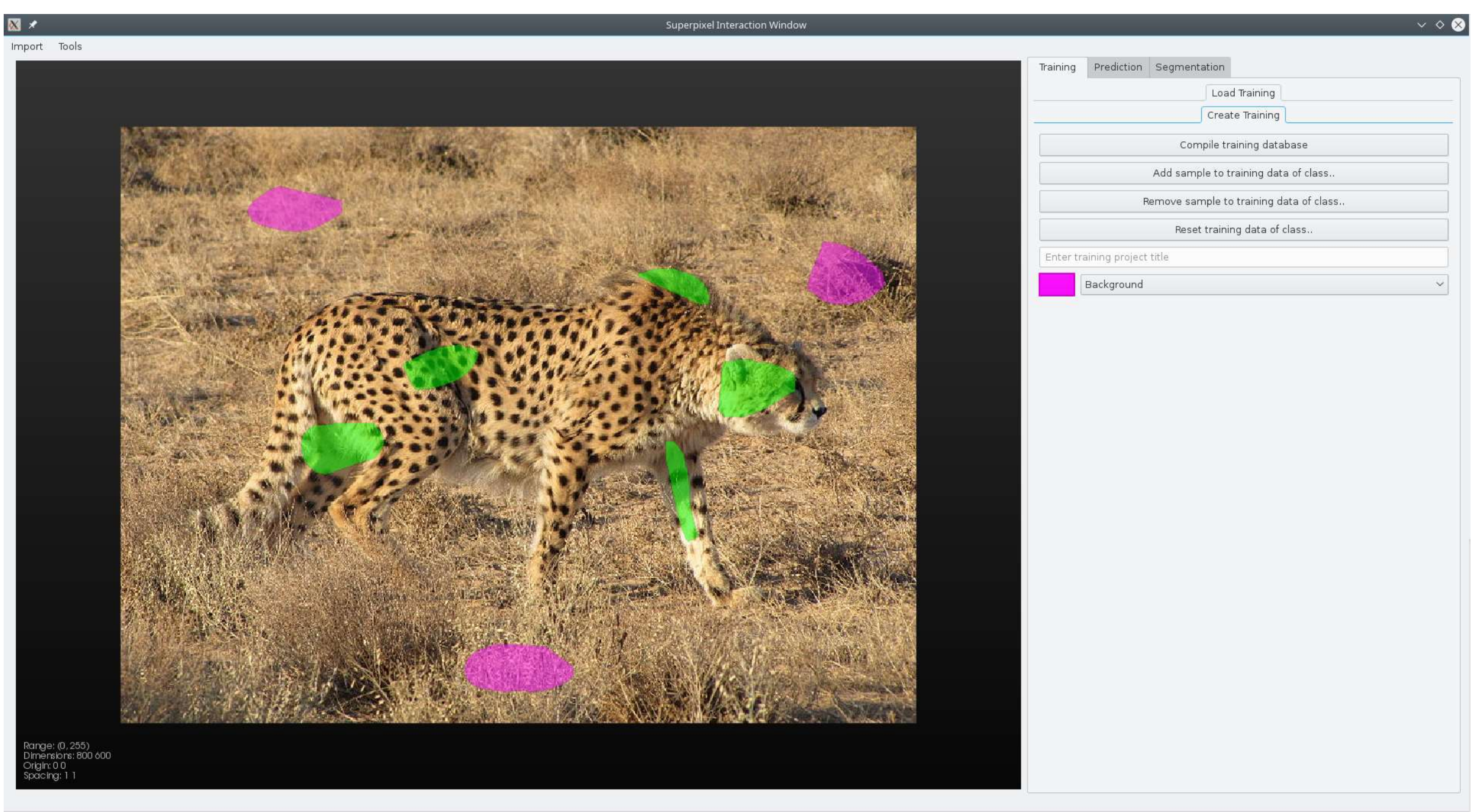

*‘Superpixel Project Viewer’ with ‘Create Training’-GUI*

***Required Preparations:***

A superpixel project (created with ‘Superpixel Preparation’ pipeline).

***Procedure:***

1. Select 'ImageProcessing' tab.
2. Select 'Superpixel Segmentation' in the pipeline selection menu.
3. Set 'Superpixel project directory' to the project folder containing the superpixel project.
4. Click 'Open Superpixel Project' button. The 'Superpixel Project Viewer' will be opened, displaying your image dataset, overlayed with training data in case you already have a training project.
5. In the menu on the right of the 'Superpixel Project Viewer' select the 'Training' tab.
6. In the 'Training' tab select the 'Create Training Project' sub-tab.
7. Under 'Class:' you can select the training class to be edited and an associated color.
8. You can edit training data of the currently selected training class by drawing the outline of a training sample region directly into the image:
    a. Drawing a training sample region:
        - Holding down the left-mouse button allows for precise drawing of the training sample region's outline. When you release the left mouse button a straight line between the point where the button was released and the current mouse position will be proposed, which will be added to the outline if you left-click again. Finalize the outline by right-clicking. You don't have to close the outline manually, a straight line will be added between its first and last point automatically when you proceed with the next step.
        - If you are not satisfied with the current training sample region, you can delete it by clicking on the button 'Clear sample' or by using the shortcut Alt-c.
        - Also, you can retroactively (after finalization via right click) refine the outline. Hovering with the mouse cursor over the outline will show small rings around certain points. You can either left click (and hold down) on such a point and move it, or press Delete key on the keyboard to remove it. If you left click on a point of the outline without a small ring, a new point is added to the outline, which in turn can be moved.
    b. Applying a training sample region to the training data
        - If you are satisfied with your training sample region, click 'Add sample to training data' or use the shortcut Alt-a and it will be added to the training data for the class selected under 'Class:'.

- If you want to correct existing training data of the currently selected class, you can remove the training sample region from the pre-existing training data by clicking 'Remove sample from training data' or using the shortcut Alt-r. One can think of it as a subtraction operation where the training sample region is subtracted from the pre-existing training region.

9. If you have added training data to both classes click 'Compile training database'. A new training project will be created, containing a training database consisting of the superpixels within the training regions.
10. All available training projects will be accessible in the 'Load Training Project' tab. The drop-down menu shows the currently selected training project (by default the one created last). By clicking on the drop-down menu, all available training projects will be displayed. If you change to a different training project, the training data overlay and the information table under the drop-down menu will be updated accordingly.

***Note:***

- Training regions can be fairly small (as in the cheetah example). Don't worry that your training data is insufficient. You can always come back and add more. Initial predictions will guide further training data refinement (more on this in the *Workflow* section).
- Add representative regions to the training data for background and foreground classes. These samples should contain the visual variability each class exhibits. (E.g. in the cheetah example, the background class training regions should at least cover blurry gras (upper left corner), sharp scrub (lower middle) and a dark region (upper right)).
- In case you are working on a 3D dataset, a slider will be displayed in the image viewer area, which can be used to scroll through the dataset (alternatively, use ↑ and ↓ keys). While training regions are drawn in 2D per slice, the 3D supervoxels that intersect those regions will be used for the training database.

***Estimated execution time:***

On a six core i7-5930k CPU with 3.50GHz and 64GB RAM for an 5000×5000 pixel image and a superpixel size of 10×10 pixel (~250k superpixel) and R/G/B gradient magnitude and R/G/B histogram features activated training database compilation took <10 sec.

## Import Training Data

***Required Preparations:***

Either binary mask(s) for the current dataset or a valid training project of a comparable dataset (same type of image, e.g. magnification, imaged object, ..; same colorization, e.g. used markers; similar visual appearance, e.g. signal-to-noise ratio) with the same feature-setup.

***Procedure:***

1. Select 'ImageProcessing' tab.
2. Select 'Superpixel Segmentation' in the pipeline selection menu.
3. Set 'Superpixel project directory' to the project folder containing the superpixel project.
4. Click 'Open Superpixel Project' button. The 'Superpixel Project Viewer' will be opened, displaying your image dataset, overlayed with training data in case you already have a training project.
5. In the menu on the right of the 'Superpixel Project Viewer' select the 'Training' tab.
6. In the 'Training' tab select the 'Create Training Project' sub-tab.
7. If you have a binary mask for the dataset at hand (e.g. you have ground-truth for it, or training data from another superpixel project, based on the same base image), you can import it, by using 'Import' -> 'Import training mask' in the menu bar:
    a. The binary mask image has to be a 8-bit image of same dimensionality and size as the original image. Background/foreground values are expected to be 0/255. A dialog will guide you to determine, to which training class the foreground regions of the imported mask will be added.
    b. If you have added training data to both classes, either by importing masks or by drawing click 'Compile training database'. A new training project will be created, containing a training database consisting of the superpixels within the training regions. It will be automatically selected and displayed in the tab 'Load Training Project'.
8. Alternatively, you can directly import a training database from another superpixel project. Select 'Import' -> 'Import training database' in the menu bar:
    a. In the following file dialog, navigate to another superpixel project, which should be based on the same type of dataset, and select a training project that led to a satisfying prediction. High visual similarity between the two datasets is key to successful reusability of the training database.
    b. Since the imported training project contains a training database already, you can now proceed directly with the 'Create Prediction' section. In case prediction

results using imported training database are not satisfying, you can create new training data as described in section 'Create Training Data' while setting the 'Add to imported training database' dropdown menu to the imported training project. The new training regions will then be added to the imported training database, thereby creating a training database containing samples from different images.

***Estimated execution time:***

No processing here.

## Create Prediction

***Example:***

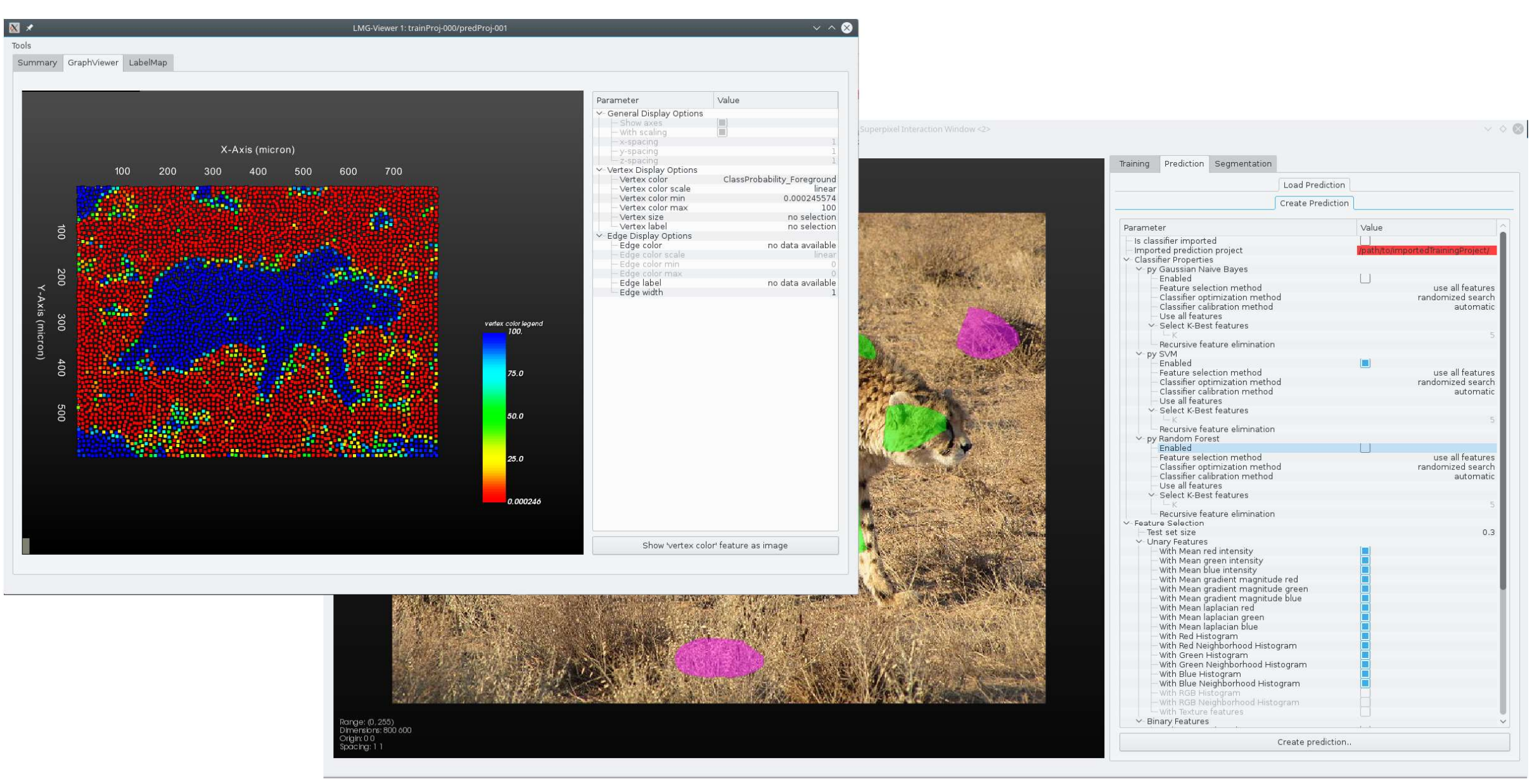

*'Create Prediction Project'-GUI and 'Prediction Project Viewer'*

***Required Preparations:***

A valid training project.

***Procedure:***

1. Select 'ImageProcessing' tab.
2. Select 'Superpixel Segmentation' in the pipeline selection menu.
3. Set 'Superpixel project directory' to the project folder containing the superpixel project.
4. Click 'Open Superpixel Project' button., which opens up the 'Superpixel Project Viewer', displaying your image dataset, overlayed with training data of the latest training project and a 'Prediction Project Viewer', showing a Prediction Project, in case one already exists.

5. In the menu on the right of the 'Superpixel Project Viewer' select the 'Prediction' tab and within it the 'Create Prediction Project' sub-tab.
6. Select under 'Classifier Properties' the classifier(s) you want to use for learning and prediction. You can choose between Support Vector Machine (SVM) and Random Forest (RF). To enable a classifier make sure 'Enabled' under the respective classifier entry is set to true.
7. You may change processing properties for the classifiers. In both cases classifier optimization can be en/disabled and the calibration method can be chosen. Usually the default values can be kept.
8. Select the features the classifier is allowed to use for learning class membership of supervoxels. By default, all available features are selected. Usually it is not beneficial to use less features. Features not added during superpixel project creation will be grayed out and not selectable.
9. Click 'Create Prediction' to create a new prediction project. Once (learning and) prediction is done a new window, the 'Prediction Project Viewer', opens, showing the results. For a brief explanation of it see the following *Note* section.
10. All available prediction projects that are based on the currently selected training project will be accessible in the 'Superpixel Project Viewer's 'Prediction' tab under the 'Load Prediction Project' sub-tab. The drop-down menu shows the currently selected prediction project (by default the one created last). By clicking on the drop-down menu, all available prediction projects will be displayed. If you change the drop down menu to a different prediction project, the information table will be updated accordingly. You can open it in a separate 'Prediction Project Viewer' by clicking the 'Open Prediction Project' button.

***Note:***

- Alternatively, you can directly import a trained classifier from another superpixel project. Select 'Import' -> 'Import classifier' in the menu bar of the 'Superpixel Project Viewer'. In the following file dialog, navigate to another superpixel project, which should be based on the same type of dataset, and select a prediction project with a satisfying result. The current superpixel project has to have the same features added, than the ones, the classifier was trained with. High visual similarity between the two datasets is key to successful reusability of the trained classifier on another dataset.
- The 'Prediction Project Viewer':
    - The 'Summary' tab gives an overview of basic information about the dataset and about settings of the superpixel and training project on which the current prediction project is based. Additionally, information concerning the prediction project (e.g. used classifier, its parametrization, used features) and the

performance of the classifier (e.g. Average precision and Brier score) are presented.

- The 'Graph Viewer' tab shows the superpixel graph, colored for probability of belonging to the foreground class.
- The 'Image Viewer' tab shows the original image dataset, and optionally overlays (e.g. segmentations).

***Estimated execution time:***

On a six core i7-5930k CPU with 3.50GHz and 64GB RAM for a 5000×5000 pixel image and a superpixel size of 10×10 pixel (~250k superpixel) and R/G/B gradient magnitude and R/G/B histogram features activated learning and predicting using a random forest took <20 sec.

## Create Segmentation

***Example:***

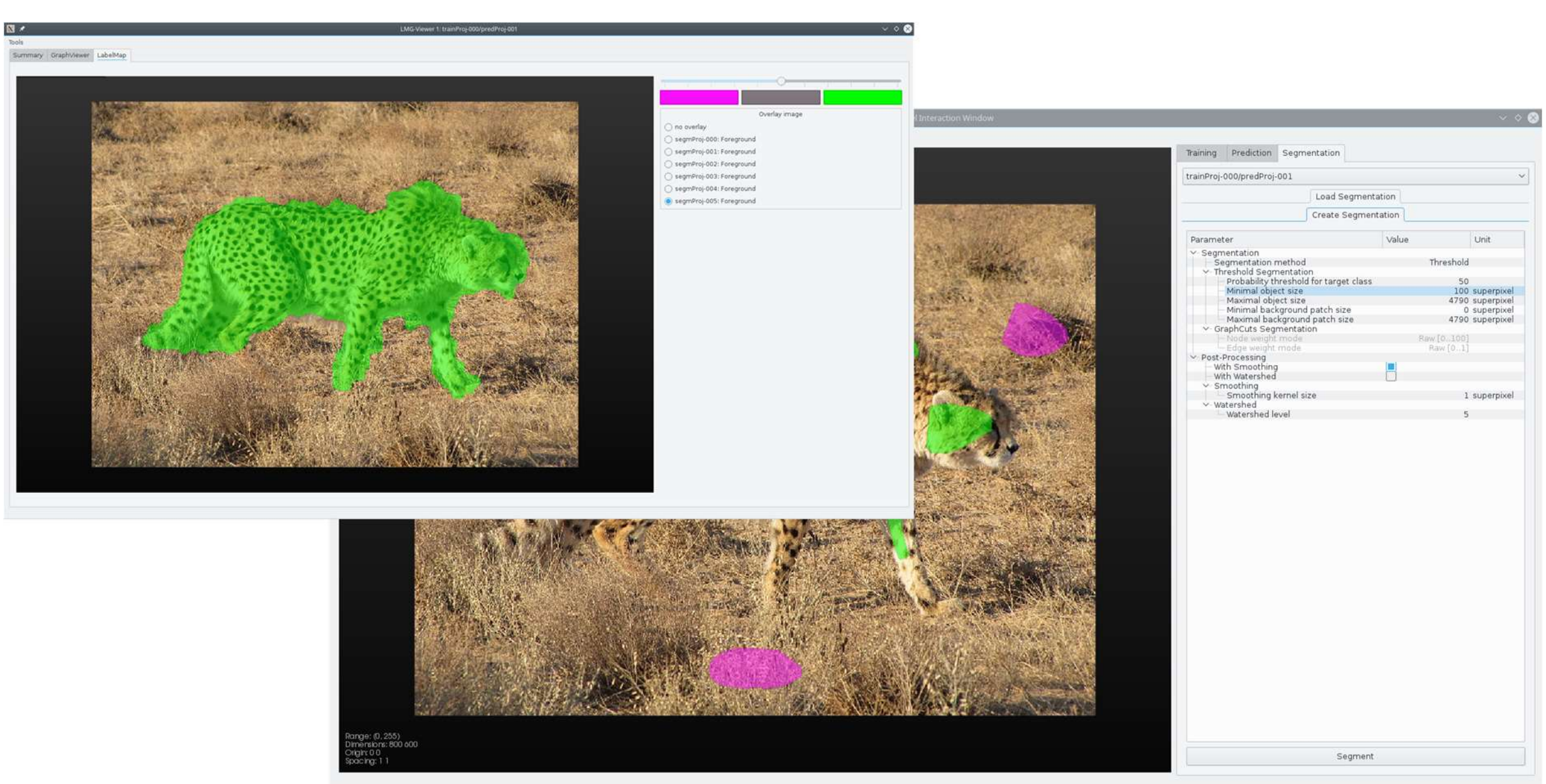

*'Create Segmentation Project'-GUI and 'Prediction Project Viewer'*

***Required Preparations:***

A valid prediction project.

***Procedure:***

1. Select 'ImageProcessing' tab.
2. Select 'Superpixel Segmentation' in the pipeline selection menu.
3. Set 'Superpixel project directory' to the project folder containing the superpixel project.

4. Click ‘Open Superpixel Project’ button., which opens up the ‘Superpixel Project Viewer’, displaying your image dataset, overlayed with training data of the latest training project and a ‘Prediction Project Viewer’ showing the latest prediction project for the selected training project.
5. In the menu on the right of the ‘Superpixel Project Viewer’ select the ‘Segmentation’ tab and within it the ‘Create Segmentation Project’ sub-tab.
6. The drop-down menu ‘Select Prediction Project Viewer’ lists all currently opened prediction projects, each shown in its own ‘Prediction Project Viewer’. Select the prediction project for which you want to create a segmentation.
7. You may specify the ‘minimal’ / ‘maximal foreground object size’ values. Foreground objects in the segmentation image that are smaller / larger will be converted to background objects.
8. Accordingly, you may specify the ‘minimal’ / ‘maximal background object size’ values. Isolated background objects in the segmentation image that are smaller / larger will be converted to foreground objects.
9. In case you want to smooth the segmentation boundaries, you have to enable the ‘With Smoothing’ option, and set an appropriate ‘Smoothing kernel size’ (given as multiples of average superpixel size). Smoothing will close holes within foreground objects and remove isolated foreground objects (smaller than the specified kernel size) and grind of corners of foreground objects. As a rule of thumb, the kernel size should be smaller than the smallest features that should be preserved (e.g. in the shown example the legs of the cheetah are approximately three superpixels wide; a smoothing kernel size of three and larger will remove the legs, whereas smaller values will preserve them).
10. In case you want to separate foreground agglomerated object clusters into individual objects, you have to enable the ‘With Watershed’ option. This method assumes roundish foreground object forms and might be used to e.g. split connected nuclei into individual objects. The sensitivity of the approach is regulated by the ‘Watershed level’. Smaller values will lead to more splits, higher values to less splits and thus resulting individual objects. This method is not able to correctly separate objects of complex shape (e.g. split two connected cheetahs).
11. Click ‘Create Segmentation’ to create a new segmentation project. Once segmentation is done it will be shown as an overlay in the ‘Image Viewer’ tab of the ‘Prediction Project Viewer’ showing the corresponding prediction project. In the directory corresponding to the segmentation project, you will find a file ending on ‘Foreground_bin.tiff’, which contains the binary mask (0=background, 255=foreground) of the segmented objects. The files ending on ‘Foreground_contour.tiff’ and ‘Foreground_overlay.tiff’ are additional visualizations of the segmentation together with the original data for quick evaluation of its quality.

12. All available segmentation projects that are based on the prediction project, currently selected with the '' drop-down list, will be accessible in the 'Load Segmentation Project' page. The drop-down menu shows the currently selected segmentation project (by default the one created last). By clicking on the drop-down menu, all available segmentation projects will be displayed. If you change the drop down menu to a different segmentation project, the information table will be updated accordingly. You can open it in the 'Prediction Project Viewer' corresponding to its prediction project by clicking the Open Segmentation Project:' button.

***Estimated execution time:***

On a six core i7-5930k CPU with 3.50GHz and 64GB RAM for a 5000×5000 pixel image and a superpixel size of 10×10 pixel (~250k superpixel) and R/G/B gradient magnitude and R/G/B histogram features activated it took ~1.3 min.

# Pipeline-based image processing

## Workflow

To ensure segmentation and quantification of results that are of high and reproducible quality, it is necessary to follow an iterative procedure.

First, all channels of a dataset that are meant to be used for segmentation have to be thoroughly examined and assessed for quality. It is imperative that structures to be segmented are clearly visible and are set apart from the background throughout the whole dataset. If a lack of staining or high background noise compromises a considerably large part of the dataset, then the dataset may not be suitable for quantification. If the dataset is considered to be of sufficient quality for quantification, it is recommended to use the default parameter set of a processing pipeline to produce an initial segmentation. After execution, the quality of the result has to be assessed. For this purpose, each pipeline produces a file ending with the suffix '_overlay', which is an overlay of the segmented structures over the raw data. In addition, the segmentation is saved as a binary mask with the file ending '_bin', which may then be used as input for other pipelines (e.g. analysis pipelines). It is recommended that the overlay image is used for quality assessment. Besides the obvious quality criteria such as alignment of segmentations to underlying structures, pipeline-specific quality criteria are noted in the corresponding pipeline section. If a segmentation result is found deficient, the problematic filter in the pipeline has to be identified. For this purpose, each filter within a pipeline produces an intermediate result, which is named according to the filter. Based on these intermediate results, filter parameters have to be adjusted to improve segmentation quality. Afterwards, the pipeline has to be executed again, and the quality assessment and parameter adjustment procedure is iterated. It is advisable to move the results obtained from earlier attempts into a subdirectory in order to examine the effect of the parameter adjustments. ***Otherwise, the results of earlier attempts will be overwritten.***

The described procedure may have to be repeated several times depending on the image quality and experience of the user.

## Required data

1. Digitized image stacks in TIFF format prepared according to the staining protocols published in [1] comprising DAPI, DPPIV/CD26, GS and DMs channels, imaged using the 60× objective at a voxel resolution of 0.207 µm × 0.207 µm × 0.54 µm and 1,024 pxl × 1,024 pxl in $x$–$y$ dimension. This set-up shows a fraction of one hepatic lobule.

***We provide two pre-processed test datasets including all segmentation results on our website. They can be used for evaluation of the software. For a detailed description on how to produce high quality segmentation results with TiQuant using these test datasets please refer to to the section 'Recipe for test dataset'.***

2. Imaging setups meant for the analysis of a complete lobule should be imaged with a 20× objective at a voxel resolution of 0.621 µm × 0.621 µm × 0.54 µm and 1,024 pxl × 1,024 pxl.

# Pre-processing algorithms

## Preparatory steps

1. Open the image stack with image processing software (e.g. ImageJ).
2. Split the channels.
3. Convert each channel to grayscale, 8-bit pixel precision and save as a tiff file.

## Contrast limited adaptive histogram equalization (CLAHE)

***Example:***

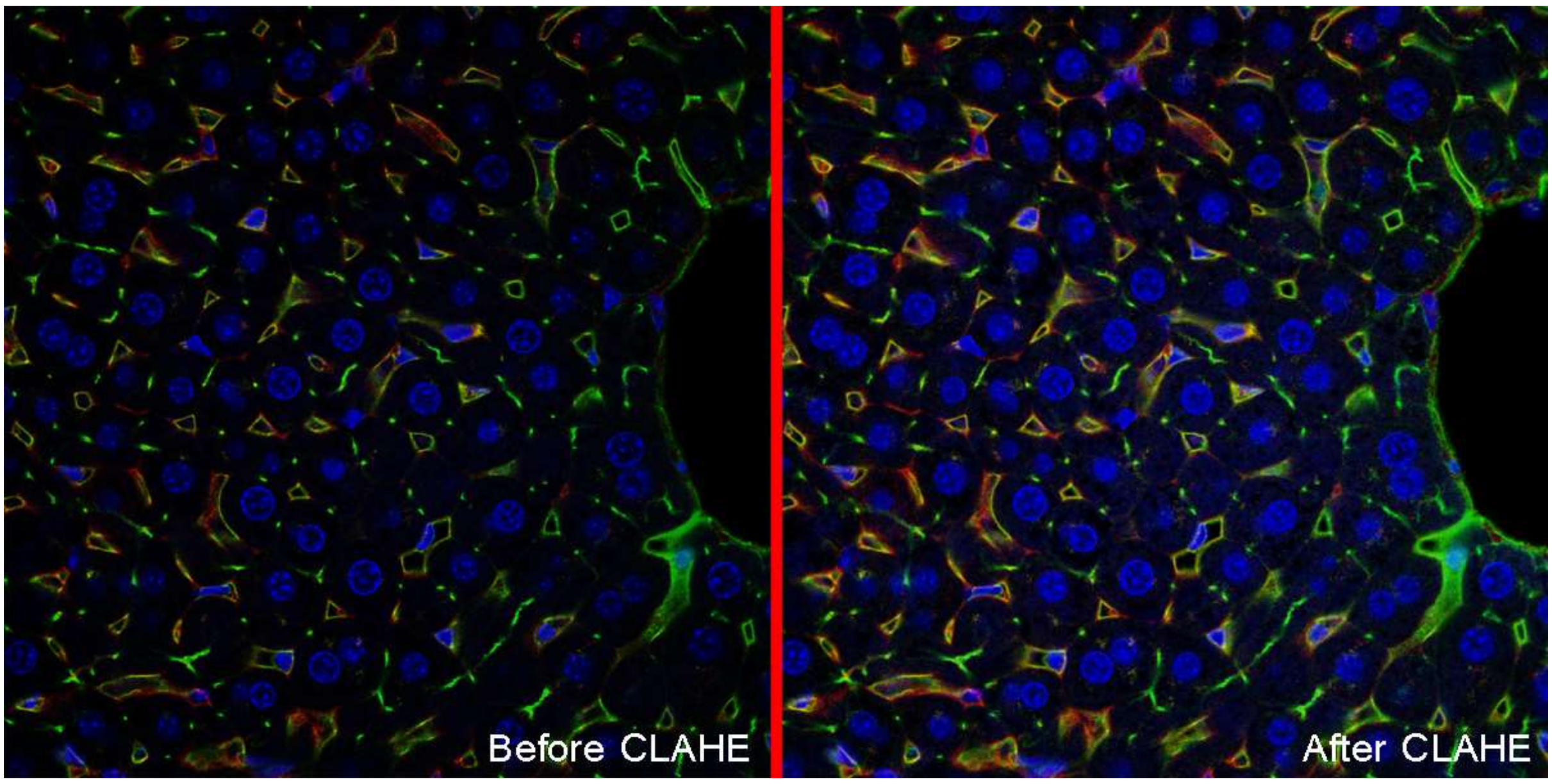

***Required Preparations:***

A three-dimensional 8-bit grayscale image saved in TIFF format is needed.

***Procedure:***

In most cases, image stack quality in terms of brightness and contrast can be improved by the application of an algorithm for contrast and brightness equalization and enhancement. The following steps should be applied to each of the channels that will be used for further segmentation steps (DAPI, DPPIV, DMs):

1. Select 'ImageProcessing' tab.
2. Select the pipeline 'CLAHE'.
3. Specify the file location of the channel that has to be processed under 'Image filename'..
4. Use the default 'CLAHE Parameters' in an image stack set-up as described above:

a. ‘Histogram Window Size’ specifies the size of the window in 2D, respectively, the cuboid in 3D, in pixel/voxel for which an intensity histogram is compiled. The size should be larger than half the size of the features that are to be amplified. The larger the window, the longer the runtime.
b. ‘Step Size’ specifies the rate in pixel/voxel at which a new histogram window is calculated along each axis.
c. ‘Clip Level’ regulates the strength with which features are amplified. The parameter range is 0–1, where 1 means that the algorithm behaves similarly to the adaptive histogram equalization (AHE) and thus produces the strongest amplification. Lower values will limit unwanted noise amplification. Values of 0.1 or smaller are reasonable for many microscopy image set-ups, depending on the amount of background noise and intensity of features.

5. Once you are finished with configuration of the pipeline click the ‘Start Pipeline’ button. After execution, a file is created in the same folder with the same name as the input file with the suffix ‘_clahe’.

***Estimated execution time:***
On a i7-2600k CPU with 3.40GHz and 16GB RAM approximately 3 hours.

## Crop

***Required Preparations:***

A three-dimensional 8-bit grayscale image saved in TIFF format is needed.

***Procedure:***
In most cases, several first and last *z*-slides contain imaging artefacts (e.g. out-of-focus blur) and have to be discarded. To do so:

1. Open all channels with image processing software and identify the first and last *z*-slide that has no distortions in any channel.
2. Select ‘ImageProcessing’ tab.
3. Select the pipeline ‘Crop’.
4. Specify the file location of the pre-processed channel under ‘Image filename’. This pipeline processes one channel at a time.
5. Set the dimensionality of the data under ‘Data dimensionality’.
6. Under ‘RegionOfInterest’ enter the number of the first and last *z*-slide of sufficient quality in the fields ‘*z* start’ and ‘*z* end’. If image stacks differ in maximal *x* and *y* dimension from the default value, fill in the used dimension in ‘*x* end’ and ‘*y* end’.

7. Once you are finished with configuration of the pipeline click the 'Start Pipeline' button. After execution, a file will be created in the same folder with the same name as the input file with the suffix '_cut'.

***Note:***

Crop each channel of an image stack using the same x, y and z bounds.

***Troubleshooting:***

Normally, it is not necessary to apply CLAHE to the GS channel. However, if the GS channel has to be later used for vein segmentation, then it must be cropped.

***Estimated execution time:***

On a i7-2600k CPU with 3.40GHz and 16GB RAM less than 1 minute.

# Segmentation

## Necrotic region segmentation

***Required Preparations:***

The DPPIV and DMs channels (optionally pre-processed) as three-dimensional 8-bit grayscale images saved in TIFF format are needed.

***Procedure:***

1. Select 'ImageProcessing' tab.
2. Select 'Segment Necrotic Region' in the pipeline selection menu.
3. Specify file locations of pre-processed DPPIV and DMs channels. The DPPIV channel will be used for result visualization only.
4. Choose the 'Threshold mode'. The sample region size of the 'Adaptive Otsu Threshold' should be at least half the diameter of the largest necrotic region along the respective axis. The number of samples regulates at how many points within the dataset a sample region is constructed, and an individual Otsu Threshold is calculated. More samples increase the processing time. The larger the sampling region, the fewer sampling points are needed. An alternative to the 'Adaptive Otsu Threshold' is the 'Manual Threshold' option, where a threshold for the whole image stack can be manually specified.
5. The first erosion-dilation pair is for noise removal and closing of small cavities within the necrotic region. The dilation kernel size should be roughly twice the size of the erosion kernel size. The kernel sizes depend largely on the level of noise within the dataset.
6. The second erosion-dilation pair is for the removal of sinusoidal structures. Therefore, kernel size of the erosion operator should be slightly larger than the sinusoidal radius (in pixel). The kernel size of the dilation operator should be slightly larger than the erosion kernel to close the remaining cavities within the necrotic region.
7. Remove small isolated objects. In the preceding step, most of the sinusoidal system was removed. However, the complete structure may not have been removed by the erosion operator if very thick sinusoids and branching structures were present. Therefore, in this step, all objects below a certain volume will be removed. Adjust the volume, in case of very small necrotic regions (e.g. at dataset borders).
8. Once you are finished with configuration of the pipeline click the 'Start Pipeline' button. After execution using the DPPIV and DMs channel and files beginning with the prefix 'necroticRegion', the segmentation is saved as a binary mask and two overlay images.

***Note:***

The boundaries of the segmented necrotic region should match the image data, as they will be used for distance measures. Holes within the segmented necrotic region are tolerable and sometimes inevitable due to the lack of DMs signal.

***Estimated execution time:***

On a i7-2600k CPU with 3.40GHz and 16GB RAM less than 1 hour.

## Vein segmentation

***Required Preparations:***

The DPPIV, DMs and (optionally) DAPI channels (optionally pre-processed) as three-dimensional 8-bit grayscale images or an RGB image saved in TIFF format are needed.

***Procedure:***

1. Select 'ImageProcessing' tab.
2. Select 'Segment Veins' in the pipeline selection menu.
3. Set 'Data dimensionality' to the dimension of the image and 'Image type' depending on whether you have a RGB image or several gray-scale images.
4. Specify file locations of the (optionally pre-processed) DPPIV, DMs and (optionally) DAPI channels, respectively of the 'RGB image'
5. Define the 'Thresholds' -> 'Background Color' for the three channels of your RGB image, respectively for the DPPIV, DMs, DAPI images. In the context of vein segmentation this relates to the *color* of the vein lumen. In case of confocal micrographs vein lumen is typically characterized by an absence of signal, therefore the background color would be (nearly) black. For whole-slide scans the background typically is illuminated, thus the background color would be some shade of white.
6. Define under 'Thresholds' for each of the channels a threshold, which is additive (in regards to the specified background color) when having dark background and subtractive when having light background.
7. The opening radius is determined by the radius of the lumen of the largest attached blood vessel that should not be part of the vein segmentation.
8. The closing radius is determined by the radius of the largest object in the vein lumen. The signals of macrophages, erythrocytes or other cells remaining in the lumen of the vein during imaging will thereby not compromise segmentation.

9. Using the 'Minimal vein size' parameter veins smaller than that will be removed from the segmentation, helping to remove remaining artificial non-vein structures / veins cut off at dataset borders.
10. Select 'Seed Point Selection' below the parameter table to open the seed point selection window (see *Critical step* section for more details on it).
    a. Place a marker in the lumen of each central vein.
    b. Place a marker in the lumen of each portal vein.
    c. Click 'Finish Seed Point Selection'. The seed points will be saved to a file, the selection window will close, and the pipeline is ready to be started.
11. Once you are finished with configuration of the pipeline click the 'Start Pipeline' button. After execution, the segmentation of central and portal veins is saved as a binary mask and overlay images in files beginning with the prefix 'vein_central' and 'vein_portal', respectively.

***Critical step:***

Seed point selection window handling:

1. Browse the stack using the up and down arrow keys or the slider. In the lower left corner, the slide number is shown.
2. In the right hand side menu bar chose which vein type to edit.
3. Colors for vein types can be changed using the two colored buttons (upper color for central veins).
4. Seed points of the currently selected type can be positioned on the image by left clicking. Placed seed points can be dragged (left click, hold down). The last placed/selected seed point can be deleted by pressing the 'Delete' key.
5. Previously generated seed points can be loaded via menu bar element 'Data' -> 'Load seed points'. The file dialog that pops up already points to the directory of the image, which is where seed point files will be stored when saved automatically. Seed points can be manually saved via 'Data' -> 'Save seed points'.
6. When seed point selection is finished click 'Finish Seed Point Selection' button, which will automatically save the seed points in two files in the root directory of the image. Seed point selection can be aborted by closing the window using the system-native close button of the window, which will prompt a dialog that asks whether to quit and with or without saving.

***Troubleshooting:***

It is crucial to identify the appropriate thresholds. Use thresholds that define the margin of the lumen as completely as possible. Two, if necessary three, channels can be used to reconstruct the veins. Holes in the endothelium that are present in all two/three threshold channels may result in erroneous reconstruction of veins, because boundary information is missing. In this case, larger opening and/or smaller closing radii may help.

***Note:***

Boundaries of the segmented veins must match with the image data, as they will be used for distance measures.

***Estimated execution time:***

On a i7-2600k CPU with 3.40GHz and 16GB RAM less than 1 hour.

## Nuclei segmentation

***Example:***

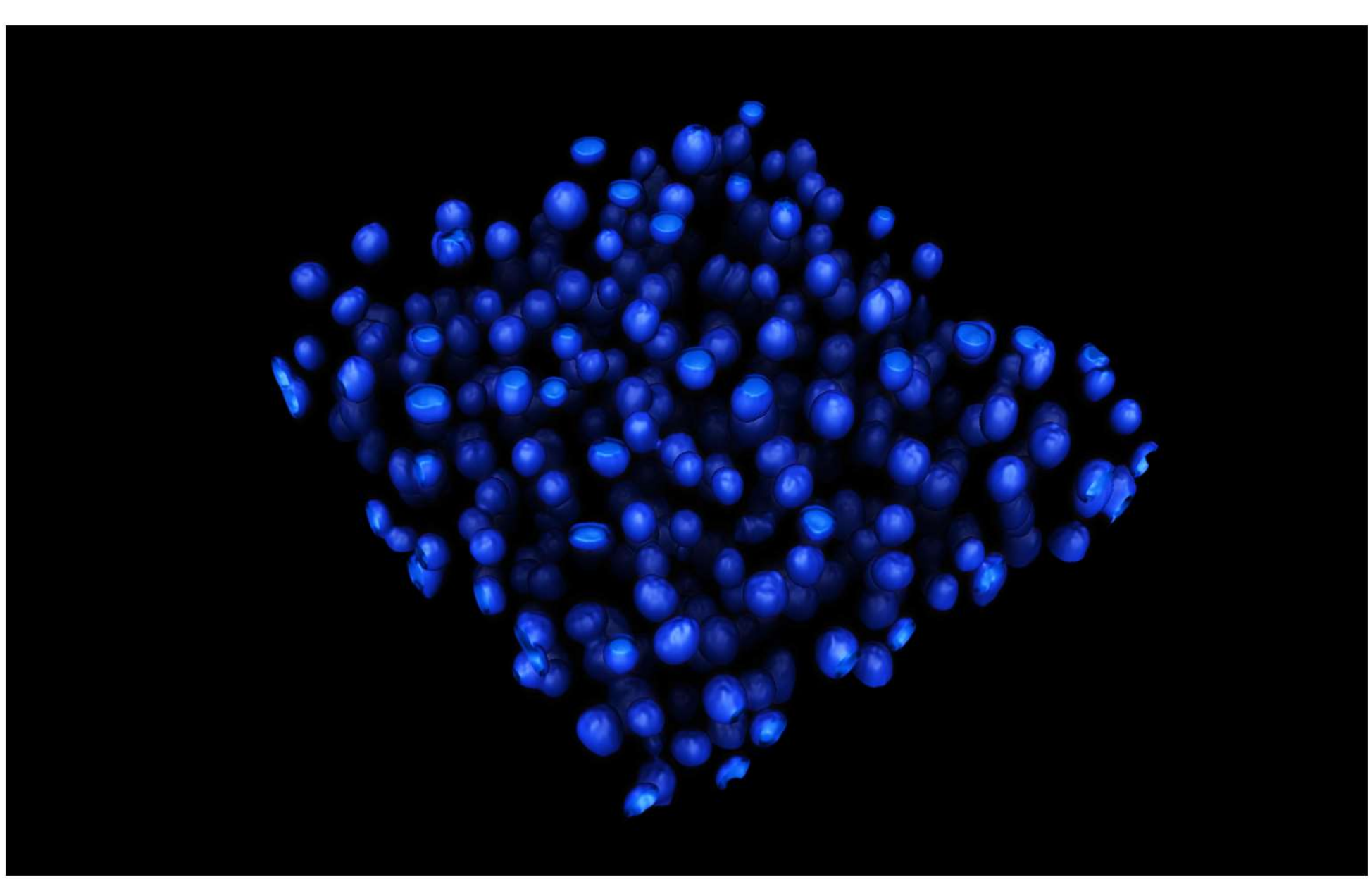

*Hepatic nuclei in a dataset (Data: TiQuant, Rendering: VolView)*

***Required Preparations:***

The DAPI channel (optionally pre-processed) as three-dimensional 8-bit grayscale image saved in TIFF format is needed.

***Procedure:***

1. Select 'ImageProcessing' tab.
2. Select 'Segment Nuclei in 60× Datasets' in the pipeline selection menu.

3. Specify file location of the pre-processed DAPI channel. Adjust the voxel spacing according to the image setup.
4. In case image data have high noise levels, increase the median filter and grayscale opening radii.
5. Choose a threshold mode. 'Adaptive Otsu Threshold' will yield superior results in most cases. In case of high noise levels or weakly stained structures of interest, the results of the Otsu method may not be satisfying quality. Try to find a 'Manual Threshold' instead.
6. Remove 'salt-and-pepper noise' using 'Inverse Hole Filling'. For high 'salt-and-pepper noise' levels, lower the 'Majority Threshold' and increase the radii. For low noise levels, the 'Majority Threshold' may be increased.
7. Close cavities in nuclei. Set the radius to a value slightly larger than the largest cavities. In case of holes that are not completely surrounded by segmented voxels, increase the 'minimal fraction of surrounding foreground' value and decrease this value if segmentation suffers from erroneously enlarged nuclei. It is recommended to use the accelerated version.
8. Close any last remaining cavities that are not completely surrounded by segmented voxels. Closing radii should be at least equal to the largest cavity radii. Too large radii may result in the artificial merging of separated objects.
9. Remove small artificial objects that are few voxels in size using the opening operator. Too large opening radii may result in the artificial separation of objects.
10. Separate the artificially agglomerated nuclei. The 'alpha' default value will be appropriate in most cases. Smaller alpha values will result in more, whereas larger alpha values will result in less separation events.
11. Remove artificial objects and group nuclei. In this step, each object will be evaluated according to its diameter. All objects with a diameter smaller than 'smallest non-hepatocyte diameter' will be dropped. Remaining objects are classified according to their diameter as non-hepatic or hepatic nuclei. In case the 'biggest non-hepatocyte diameter' exceeds the 'smallest hepatocyte diameter', nuclei with a diameter in this overlapping range will be classified according to their roundness measure, which ranges from 1 for spherical objects to 0 for infinitely elongated objects.
12. Once you are finished with configuration of the pipeline click the 'Start Pipeline' button. After execution, the segmentation of hepatic and non-hepatic nuclei is saved as binary mask and overlay images in files beginning with the prefix 'hepNuclei' and 'non-HepNuclei', respectively.

***Note:***

Segmented nuclei should contain no inner holes and must be separated and classified correctly. Separation of nearly spherical agglomerations of nuclei may fail in some cases.

***Estimated execution time:***

On a i7-2600k CPU with 3.40GHz and 16GB RAM less than 1h.

## Sinusoid and Bile Canaliculi segmentation

***Example:***

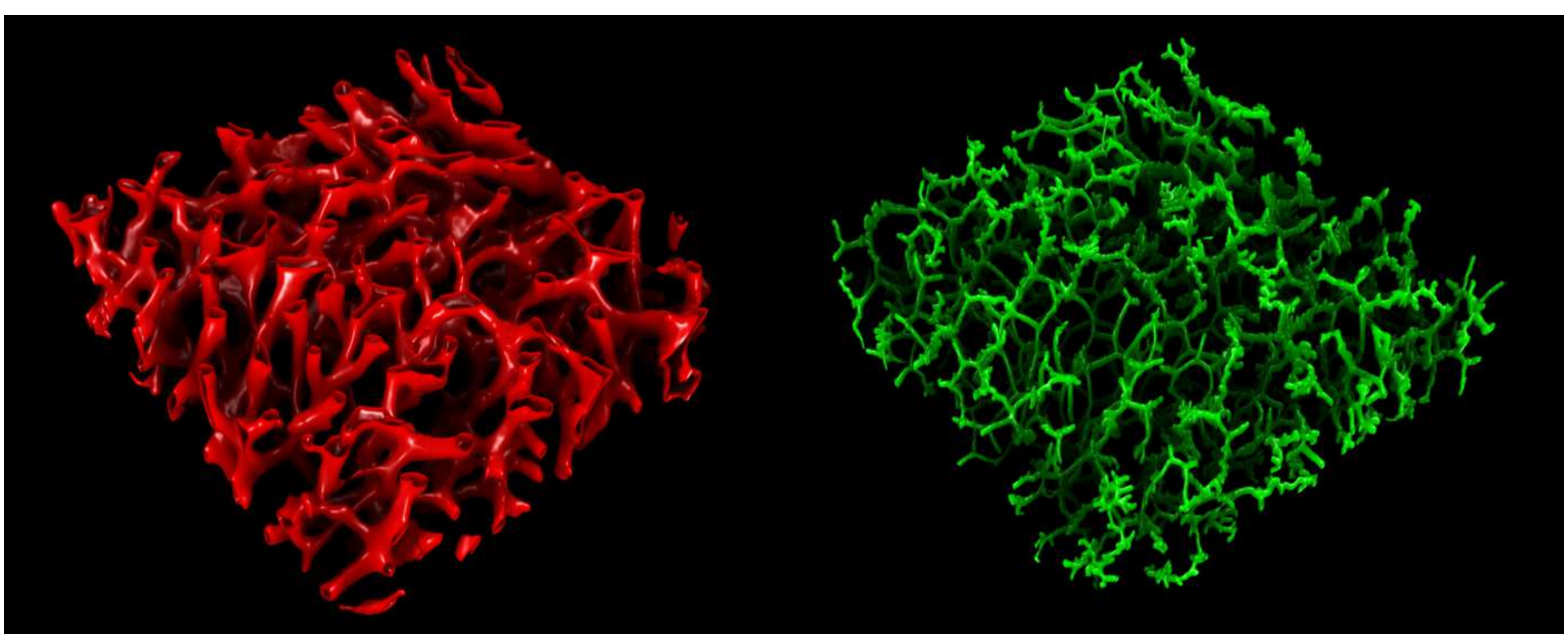

*Left: Sinusoidal blood vessels, Right: Bile network in dataset (Data: TiQuant, Rendering: VolView)*

***Required Preparations:***

The DPPIV and DMs channels (optionally pre-processed) as three-dimensional 8-bit grayscale images saved in TIFF format are needed. A segmentation of the necrotic region is necessary, in case the dataset contains one.

***Procedure:***

1. Select 'ImageProcessing' tab.
2. Select 'Segment Sinusoids + Bile Canaliculi in 60× Datasets' in the pipeline selection menu.
3. Specify file locations of the pre-processed DPPIV and DMs channels. In case the image stack contains necrotic tissue, check the 'Is there a necrotic region' box and specify the necrotic region segmentation file. Adjust voxel spacing settings if the image set-up differs from the default values.
4. Select threshold modes for segmentation of sinusoids in DPPIV and DMs channels. Recommended is the 'Adaptive Otsu Threshold'. Some datasets may require the specification of a manual threshold. A voxel needs to pass thresholds in both channels to be considered a sinusoidal voxel, which helps minimize false positives. However, locally impaired intensity of one channel may prevent correct segmentation; therefore, careful threshold tuning is mandatory.

5. Remove noise artefacts. For the removal of very small artificial objects due to 'salt-and-pepper noise' an inverse hole-filling operator is applied. Radii should not exceed 3 pixel. Majority threshold values have to be within the range of 0–10. The smaller the value, the stronger the noise reduction effect, where values smaller than 2 may also remove some boundary voxels of larger objects.
6. Fill cavities within sinusoids. It is recommended to use the accelerated (and less precise) version of the algorithm. Set the radius to a value slightly larger than the largest cavity. In case there are holes that are not completely surrounded by segmented voxels, increase the 'minimal fraction of surrounding foreground' value and decrease this value once artefactual contacts are introduced. It is recommended to use the 'accelerated' version.
7. Close the remaining cavities and discontinuities and remove small artificial objects. The closing radii should equal the largest remaining cavity radii. Opening radii should be slightly larger than radii of the largest artificial objects.
8. Remove isolated objects. All isolated objects of a volume smaller will be removed.
9. Remove noise in the DPPIV channel prior to the bile canaliculi segmentation. In case of high noise levels in the DPPIV channel, the median and greyscale opening radii might be increased to a maximum value of 2.
10. Select a threshold mode for segmentation of bile canaliculi in DPPIV channel. Recommended is the 'Adaptive Otsu Threshold'. Some datasets may require a manual threshold. Since in this step the focus is on the bile canaliculi, the method and/or value may differ from the one used in step 4.
11. Close cavities and small discontinuities. The radii of the hole-filling operator should not exceed 2. The majority value should be smaller than 4 to ensure the closing of very small gaps. Smaller majority values may result in additional boundary voxels. Make sure that branching points are still clearly defined.
12. Removal of artificial objects. There are three steps in the noise removal process. The opening may affect segmented bile canaliculi if their radius is in the range of the opening radius. In this case, skip the opening by using the default radius value of 0. Small artificial objects will be removed by inverse hole filling. Proceed as in step 5. The final step in object removal will discard all objects with a volume smaller than the specified value.
13. Once you are finished with configuration of the pipeline click the 'Start Pipeline' button. After execution, the segmentation of sinusoidal and bile canaliculi networks is saved as binary mask and overlay images in files beginning with the prefix 'sinus' and 'bile', respectively.

***Note:***

It is critical that the segmented sinusoids and bile canaliculi appear as solid structures free from inner holes. Otherwise, the network parameters will be altered. Segmentation boundaries have to match the endothelial staining in the DPPIV/DMs channel as precisely as possible. Pay special attention to the connectivity of the bile canaliculi network. Limit the number of artificial branch gaps, especially at intersections, to a minimum. Otherwise, the number of dead-end branches and intersections may be increased or decreased.

***Estimated execution time:***

On a i7-2600k CPU with 3.40GHz and 16GB RAM less than 1h.

## Network graph extraction and analysis

***Example:***

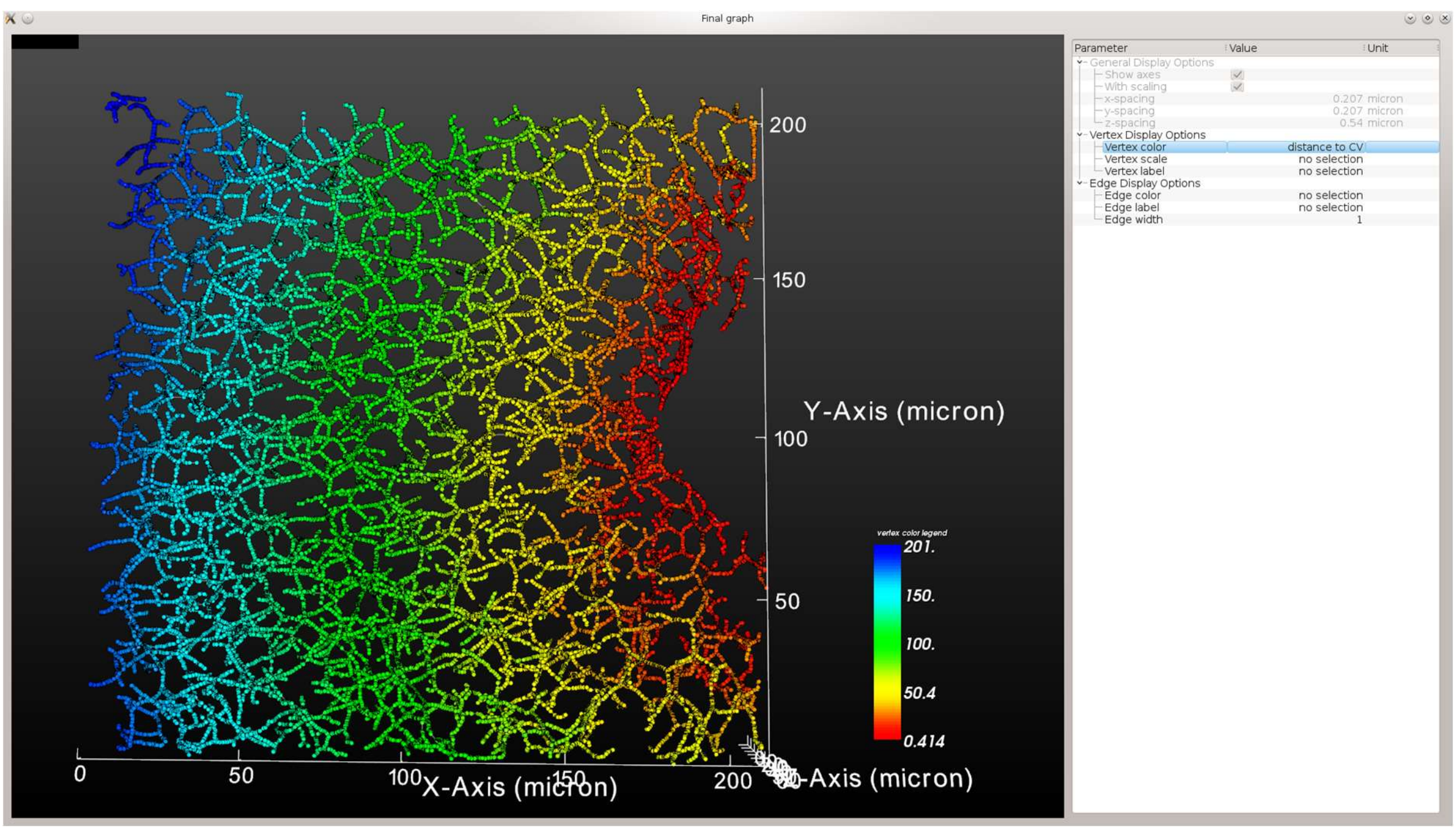

*TiQuant's graph viewer visualizing a bile canaliculi network colored according to distance to a central vein.*

***Required Preparations:***

A segmentation of the network to be analyzed is necessary. In case the network graph is not already available the network skeleton is a mandatory input. Otherwise, the graph files can be used directly to shorten computation time and prevent unnecessary recomputation of the graphs. For graph analysis additional optional inputs are segmentations of the veins (recommended, if obtained) as well as a segmentation of the necrotic region, in case the dataset contains one.

***Procedure:***

1. Select 'ImageProcessing' tab.
2. Select 'Extract and Analyze Graph' in the pipeline selection menu.
3. Choose the network type from the drop-down menu. This is a convenience option that will determine the network type-dependent default parameters.
4. Choose the type of input from the drop-down menu. The 'Skeleton Image' option will transform a skeleton image of a network into a graph representation and optionally visualize and analyse it. The 'Graph' option skips the transformation steps.
5. Depending on the input type, specify the file location of the skeleton image or the first graph file (ending on the suffix '_graph0.txt'). In case you have a "dataset file list" *.ias file (generated in the folder where segmentation files are located) you can specify this first and all necessary files will be added automatically to the interface.
6. Validate this for both network types using the following parameters:
    a. Resampling filter: The transformation of a network skeleton into a network graph produces a graph where vertices represent single voxels. Thus, graph edges have a length of single voxels. The re-sampling factor determines the fraction of vertices per branch that are kept. Intermediate vertices will be discarded. This procedure ensures, that stepping artefacts don't compromise the quantification.
    b. Remove dead-end filter: All dead-end branches of a length smaller than the threshold are considered artificial and will thus be discarded.
    c. Collapse intersection nodes filter: All intersection node paired within a distance smaller than the threshold are considered artificially separated and will thus be merged together.
    d. Geometric pruning filter: Vertices that introduce a deviation from a straight line of less than this angle are considered obsolete for maintaining the geometrical structure of the graph and will thus be deleted. Instead of this vertex and the two attached edges, a direct edge between the two other edge defining vertices is introduced. This procedure is applied to vertices with exactly two attached edges.
    e. Remove Isolated Graphs: Graphs with a total edge length shorter than this value will be removed and not considered for analysis.
7. Select whether an analysis of the network graph is desired.
8. The 'Output file prefix' will be used as a prefix for the analysis files. It is recommended to use the default value.
9. Specify the dataset name. All quantified objects will contain a reference to the dataset name in the analysis file.
10. Specify the file locations of the requested segmentation files.

11. Specify the voxel size in case it differs from the default values.
12. Once you are finished with configuration of the pipeline click the 'Start Pipeline' button. Pipeline execution graphs will be visualized, if the 'Display Options' are set accordingly. Graphs are saved in text files beginning with the prefix from point 7. Sub-graphs that are not connected are stored in individual, enumerated files.

***Estimated execution time:***

On a i7-2600k CPU with 3.40GHz and 16GB RAM less than 1 hour.

## Hepatocyte shape reconstruction

***Example:***

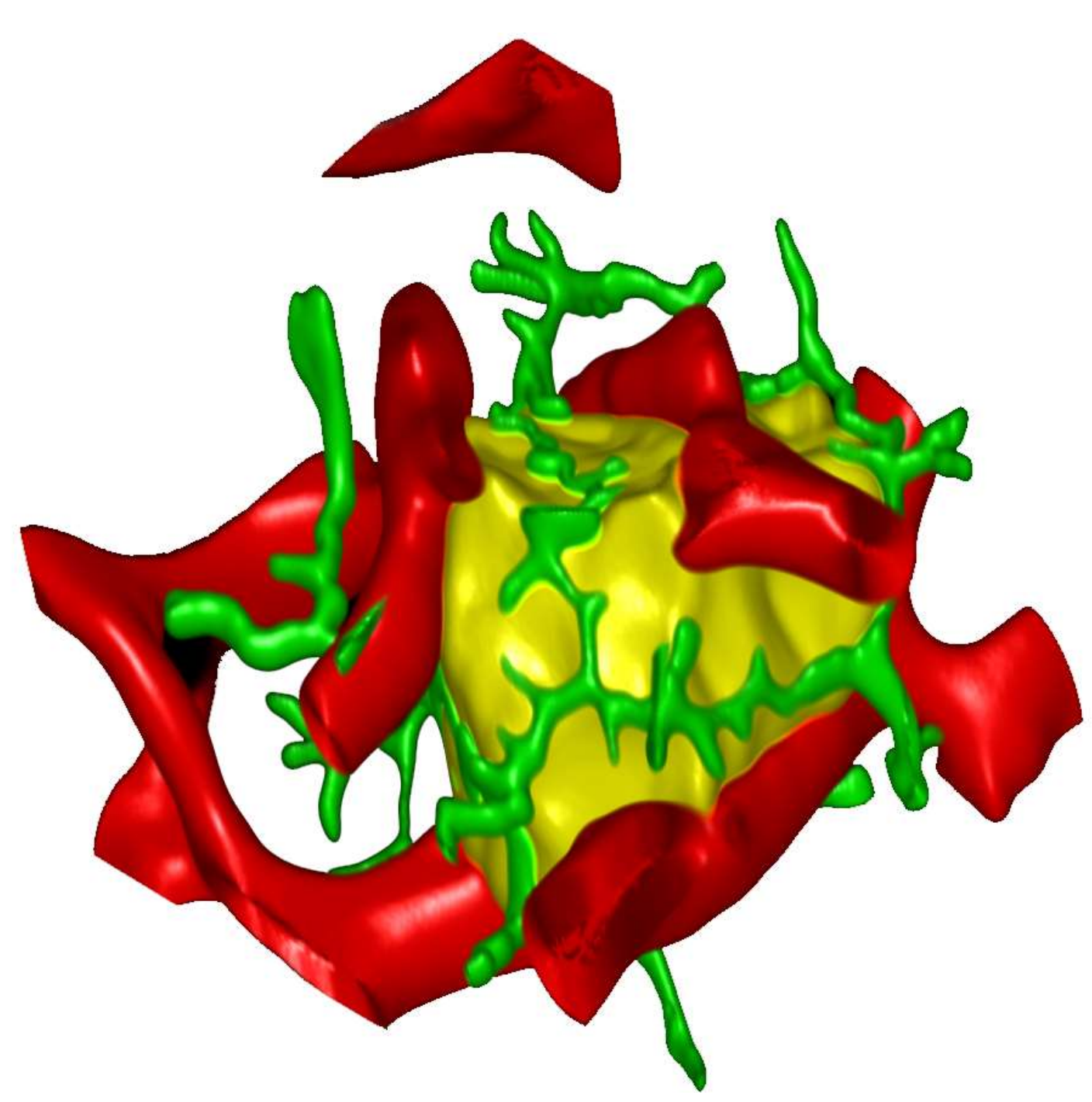

*Single hepatic cell with adjacent sinusoidal and bile networks (Data: TiQuant, Rendering: VolView)*

***Required Preparations:***

The DPPIV channel (optionally pre-processed) as three-dimensional 8-bit grayscale images saved in TIFF format is needed. Additionally, segmentations of hepatic nuclei, bile canaliculi and sinusoids are mandatory. Optional inputs are segmentations of the veins (recommended, if obtained) and a segmentation of the necrotic region, in case the dataset contains one.

***Procedure:***

1. Select 'ImageProcessing' tab.
2. Select 'Approximate Cell Shape' in the pipeline selection drop down menu.

3. Specify the location of the DPPIV channel, as well as hepatocyte, bile and sinusoid segmentation files. Central and portal vein segmentation files are optional but recommended. If there is a necrotic region in your dataset check the checkbox and specify the file. In case you have a "dataset file list" *.ias file (generated in the folder where segmentation files are located) you can specify this first and all segmentation files will be added automatically to the interface. Fill in the voxel spacing (size of a voxel) of your data.
4. It is recommended to use the default parameters in a first shape reconstruction pass. If the result is not satisfying the following parameters can be adjusted:
    a. 'Bile weight':
    The shape reconstruction makes use of already segmented objects. These objects can be classified into objects preferentially located 1) in the interior (class CI) and 2) on the 3D surface searched for (class CS). Naturally, the structures in class CS do not need to cover the entire 3D surface. We apply a Signed Maurer Distance Transformation published in [4] to obtain the Euclidean distance $d_{CI,CS}(\underline{x})$ to the nearest voxel of CI (CS) for each point $\underline{x}$ = (x,y,z) in order to compute the gradient function $g(\underline{x})$ using Equ.1:.
    $$g(\underline{x}) = \beta * d_{CI}(\underline{x}) / (\beta * d_{CI}(\underline{x}) + (1-\beta) * d_{CS}(\underline{x})) \quad (1)$$
    The bile weight parameter $\gamma (0 \leq \gamma \leq 1)$yields $\beta = 1 - \gamma$in Equ. 1 and allows balancing the influence of the two classes on the resulting gradient magnitude profile; higher values of $\gamma$increase the influence of class CS (bile and sinusoidal objects) and reduce the influence of class CI (nuclei).
    b. 'Alpha' is a parameter controlling the outcome of the watershed algorithm. Before application of watershed the gradient magnitude profile is scaled to the range 0-100. The watershed implementation we use deploys seed points to avoid oversegmentation. These seed points are all connected regions in the scaled gradient magnitude profile with values smaller than alpha. Smaller alpha values will result in more, whereas larger alpha values will result in less individual cells.
    c. Artificial cells, including cells without nucleus and cells with diameter smaller than 'minimal cell diameter' will be discarded.
5. The position filter restricts the cell reconstruction to a certain location. A pixel location has to be specified together with the number n of cells to keep. If the position filter is enabled the cell reconstruction will contain the n closest cells to the specified position only. The filter is disabled by default.
6. The save selector allows for adjustment of the log and result file names. Additionally, the save mode can be selected. Save everything will save, additionally to the end results, the intermediate results also, which will be skipped if Save only essentials is selected.

7. Once you are finished with configuration of the pipeline click the 'Start Pipeline' button.

***Note:***

At dataset borders, cell boundaries may be incorrect due to incomplete confinement information. These artifactual cells can be ignored, since only inner cells that are not in contact with dataset borders are used for quantification.

***Estimated execution time:***

On a i7-2600k CPU with 3.40GHz and 16GB RAM less than 1 hour.

## Hepatocyte analysis

***Required Preparations:***

Segmentations of cells, hepatic nuclei, bile canaliculi and sinusoids are mandatory inputs and have to be obtained before. Optional inputs are segmentations of the veins (recommended, if available).

***Procedure:***

1. Select 'ImageProcessing' tab.
2. Select 'Analyze Cells' in the pipeline selection menu.
3. Specify the dataset name. All quantified objects will hold a reference to the dataset name in the analysis file.
4. Specify the location of the cell, bile and sinusoid segmentation files. Central and portal vein segmentation files are only necessary, if available. In case you have a "dataset file list" *.ias file (generated in the folder where segmentation files are located) you can specify this first and all segmentation files will be added automatically to the interface.
5. In case of a different voxel size, correct the voxel spacing parameters.
6. Once you are finished with configuration of the pipeline click the 'Start Pipeline' button.

***Estimated execution time:***

On a i7-2600k CPU with 3.40GHz and 16GB RAM less than 1 hour.

## How to handle data with different image specifications

The effective voxel size of our datasets was 0.207 µm × 0.207 µm × 0.54 µm. The default parameter values of all pipelines are dependent on this dataset set-up.

In case the voxel size differs due to e.g. different magnification or voxel resolution, all kernel values have to be rescaled. For example, a closing operator using a kernel of size 2, 2, 1 in our set-up would use a rescaled kernel of size 4, 4, 1 in a set-up with voxel sizes of 0.414 µm × 0.414 µm

× 0.54 μm. However, this rule of thumb is only applicable when the feature appearances of all structures of interest to be segmented are similar in appearance to our setup. If this prerequisite is not met (e.g. due to lower magnification where sinusoids appear as solidly stained structures, rather than stained endothelial cells that encircle a stain-free lumen), necessary parameter adjustments might be less straightforward; and in extreme cases, the pipelines may not be applicable.

# Quantification – Parameter Legend

## Network Quantification

### Preface

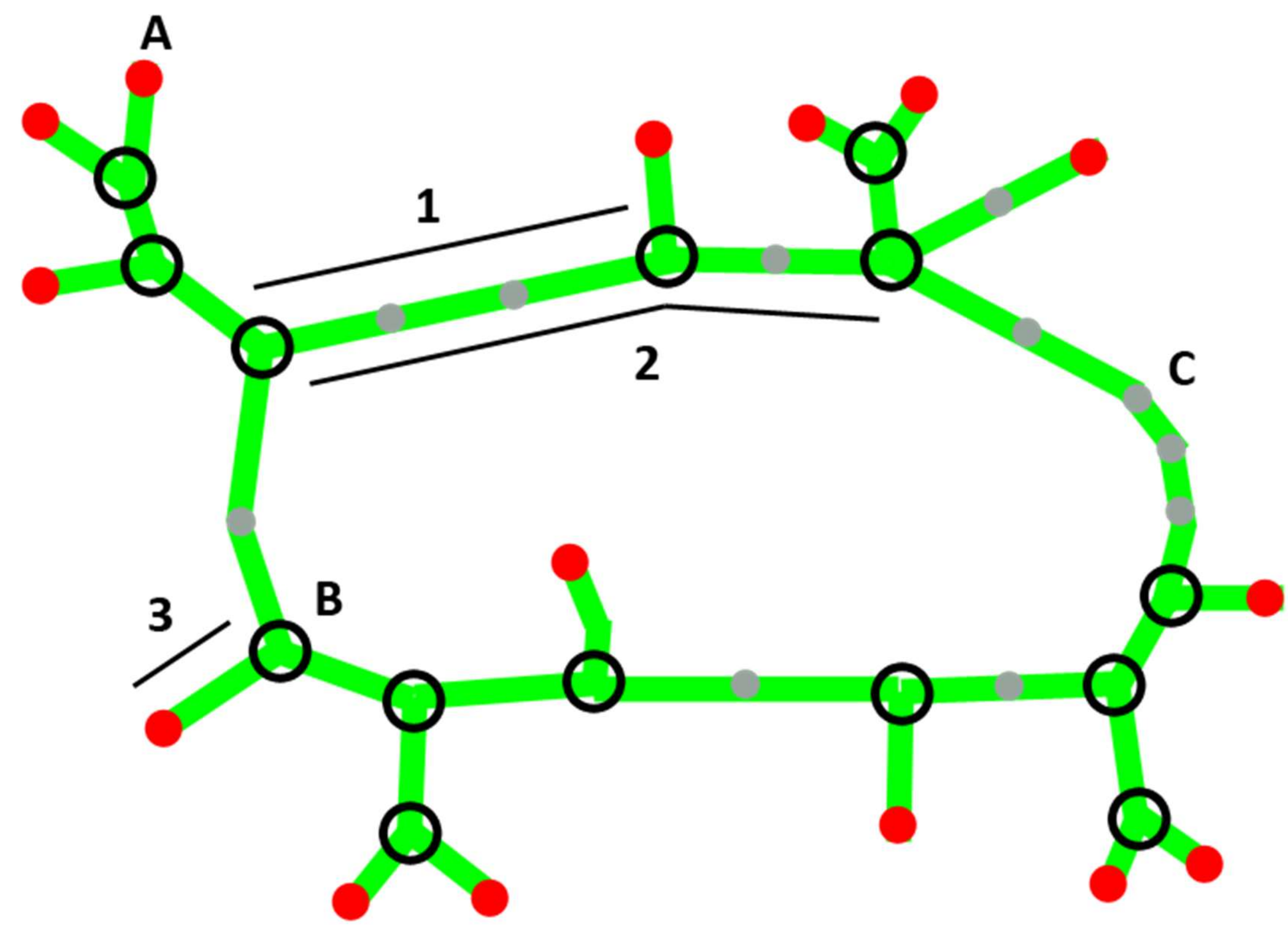

*Schematic network graph. (A) Dead end node, (B) Intersection node, (C) Intermediate node, (1) Intersection branch, (2) Second order branch, (3) Dead end branch.*

The network graph is composed of nodes and edges. The following three categories of nodes can be differentiated: dead end nodes, intersection nodes and intermediate nodes. A dead end nodes has exactly one connected edge, intermediate nodes exactly two connected edges and intersection nodes three or more connected edges. Additionally, we define three categories of branches: First-, second- and third-order branches. We differentiate between two types of first-order branches: Dead end branches that span between a dead end node and an intersection node and intersection branches, that span between two intersection nodes. Second-order branches consist of one or more intersection branches and span between two intersection nodes, that have at least three intersection branches attached to themselves. Finally, third-order branches are comprised by second-order branches, dead end branches, and sequences of intersection branches that start and end at intersection nodes with one or more than two intersection branches connected to them.

## Network Output file 1

Contains dataset-specific parameters.

First column is header. First line lists analyzed dataset names, one column per dataset.

Category explanations:

- “#dim” – Number of dimensions of dataset, e.g. 2/3 for 2D/3D
- “dimX” – Size of dataset along x-axis (in pixel)
- “dimY” – Size of dataset along y-axis (in pixel)
- “dimZ” – Size of dataset along z-axis (in pixel)
- “dataSetVolume” – Total volume of dataset (in $mm^3$)
- “veinVolume” – Vein volume in dataset (in $mm^3$)
- “effectiveVolume” – Effective tissue (dataSetVolume - veinVolume) in dataset (in $mm^3$)
- “#nodes” – Total number of nodes in network graph
- “#nodes/effVol” – Total number of nodes in network graph (per 1 $mm^3$)
- “#edges” – Total number of edges in network graph
- “length_of_network (mm)” – Total length of network graph (in mm)
- “length_of_network (mm)/effVol” – Total length of network graph (in mm) (per 1 $mm^3$)
- “avg_inscrRadius(micron)” – Mean maximal inscribed sphere radius (measured at each node) of network. Measured at each node using maximal inscribed sphere. (in micron)
- “avg_bestFitRadius(micron)” – Mean best fit radius (measured at each node) of network. Measured radius at each node in eight different directions and averaged over them. (in micron)
- “volCalc_of_network (mm^3)” – Volume of network (calculated, based on length & mean maximal inscribed sphere radius) (in $mm^3$)
- “volCalc_of_network (mm^3)/effVol” – Volume of network (calculated, based on length "& mean maximal inscribed sphere radius) (in $mm^3$ per 1 $mm^3$))
- “volCount_of_network (mm^3)” – Volume of network (counted based on number of network voxels) (in $mm^3$)
- “volCount_of_network (mm^3)/effVol” – Volume of network (counted based on number of network voxels) (in $mm^3$ per 1 $mm^3$))
- “#fOBranches” – Total number of first order branches
- “#fOBranches/effVol” – Total number of first order branches (per 1 $mm^3$)
- “avg_#edges/fOBranch” – Mean number of edges per first order branch
- “#dend_fOBranches” – Total number of dead end branches
- “#dend_fOBranches/effVol” – Total number of dead end branches (per 1 $mm^3$)

- "#isec_fOBranches" – Total number of intersection branches
- "#isec_fOBranches/effVol" – Total number of intersection branches (per 1 $mm^3$)
- "#intersec_nodes" – Total number of intersection nodes
- "#intersec_nodes/effVol" – Total number of intersection nodes (per 1 $mm^3$)
- "avg_#branches/intersec_node" – Mean number of branches per intersection node
- "var_#branches/intersec_node" – Variance of number of branches per intersection node
- "#sOBranches" – Total number of second order branches
- "#sOBranches/effVol" – Total number of second order branches (per 1 $mm^3$)
- "avg_#fOBranches/sOBranch" – Mean number of first order branches per second order branch
- "avg_fOBranch_length" – Mean length of first order branches (in micron)
- "avg_dEnd_fOBranch_length" – Mean length of dead end branches (in micron)
- "avg_isec_fOBranch_length" – Mean length of intersection branches (in micron)
- "avg_sOBranch_length" – Mean length of second order branches (in micron)
- "avg_dEnd_fOBranch_inscrRadius" – Mean maximal inscribed sphere radius of dead end branches (in micron)
- "avg_dEnd_fOBranch_bestFitRadius" – Mean best fit radius of dead end branches (in micron)
- "avg_isec_fOBranch_inscrRadius" – Mean maximal inscribed sphere radius of intersection branches (in micron)
- "avg_isec_fOBranch_bestFitRadius" – Mean best fit radius of intersection branches (in micron)
- "avg_sOBranch_inscrRadius" – Mean maximal inscribed sphere radius of second order branches (in micron)
- "avg_sOBranch_bestFitRadius" – Mean best fit radius of second order branches (in micron)
- "avg_isecBranchMinAngle" - Averaged over all intersection nodes: The smallest angle spanned by a pair of edges (considering all edge-pairs) (in °) [0..180]
- "avg_isecBranchAvgMinAngle" - Averaged over all intersection nodes: Determines for every node the minimal angle (spanned by a pair of edges, considering all other edges) and then averages over the found minimal angles (: one angle per edge) (in °) [0..180]

### Network Output file 2

Contains node-specific parameters.

First line is header. One line per network node (=vertex).

If distances are computed the position of the node is used.

Category explanations:

- “dataSet” – Dataset name in which node was found
- “#edges” – Number of edges connected to node
- “maskedDeadEnd” - True, if dead end node is considered artificial (touches dataset border or very close proximity to vein), false, if it is not considered artificial and nan, if node is not a dead end node
- “inscrRadius” – Radius of network subsegment (measured at this graph node using maximal inscribed sphere) (in micron)
- “bestFitRadius” – Radius of network subsegment (measured at this graph node by calculating radius in eight different directions and subsequent averaging) (in micron)
- “minAngle” - If more than two edges connect to the node, the smallest angle spanned by a pair of edges (considering all edge-pairs) (in °) [0..180], NAN otherwise
- “avgMinAngle” - If more than two edges connect to the node, determines for every edge the minimal angle (spanned by a pair of edges, considering all other edges) and then averages over the found minimal angles (: one angle per edge) (in °) [0..180], NAN otherwise
- “distToDatasetBorder” - Shortest distance to dataset border (in micron)
- “distToCV” – Shortest distance to any central vein pixel; NAN if no CV present
- “distToPV” – Shortest distance to any portal vein pixel; NAN if no PV present
- “minSecOrderLength” - Length of shortest second order branch originating at this node (in micron); NAN if no second order branch is originating at this node
- “maxSecOrderLength” - Length of longest second order branch originating at this node (in micron); NAN if no second order branch is originating at this node

### Network Output file 3

Contains first order branch-specific parameters.
First line is header. One line per first order branch.
If distances are computed the position of the centroid of the branch is used (considering all nodes on the branch: start + intermediate + end nodes).
If angles are computed the vector spanned by positions of start and end node is used.

Category explanations:

- “dataSet” – Dataset name in which first order branch was found

- “subGraph” - Id of sub-graph in which first order branch was found
- “id” - Unique id of first order branch
- “isIsec” – 1 if the first order branch is an intersection branch, 0 otherwise
- “isDead” – 1 if the first order branch is a dead end branch, 0 otherwise
- “connectedBranches” - Enumeration of connected first order branches (unique ids)
- “length” – Length of first order branch (in micron)
- “inscrRadius” – Average maximal inscribed sphere radius of network subsegment (measured over maximal inscribed sphere radii of all branch comprising nodes) (in micron)
- “bestFitRadius” – Average best fit radius of network subsegment (measured over best fit radii of all branch comprising nodes) (in micron)
- “distToCV” – Shortest distance to any central vein pixel (in micron); NAN if no CV present
- “distToPV” – Shortest distance to any portal vein pixel (in micron); NAN if no PV present
- “lengthInscrRadiusRatio” - Ratio of length to maximal inscribed sphere radius
- “lengthBestFitRadiusRatio” - Ratio of length to best fit radius
- “cvPathFirstOrderDegree” - First order path degree to central vein, indicates the minimal number of first order branches (first order branching nodes) needed to be traversed to reach a central vein; Nan if no CV present
- “cvPathLengthOffset” - Path length offset for central vein, indicates the length of the path that yielded the *cvPathFirstOrderDegree*; Nan if no CV present
- “pvPathFirstOrderDegree” - First order path degree to central vein, indicates the minimal number of first order branches (first order branching nodes) needed to be traversed to reach a portal vein; Nan if no PV present
- “pvPathLengthOffset” - Path length offset for portal vein, indicates the length of the path that yielded the *pvPathFirstOrderDegree*; Nan if no PV present

### Network Output file 4

Contains second order branch-specific parameters.
First line is header. One line per second order branch.

Category explanations:

- “dataSet” – Dataset name in which second order branch was found
- “subGraph” - Id of sub-graph in which second order branch was found
- “id” - Unique id of second order branch
- “length” – Length of second order branch (in micron)

- “inscrRadius” – Average maximal inscribed sphere radius of network subsegment (measured over maximal inscribed sphere radii of all branch comprising nodes) (in micron)
- “bestFitRadius” – Average best fit radius of network subsegment (measured over best fit radii of all branch comprising nodes) (in micron)
- “distToCV” – Shortest distance to any central vein pixel (in micron); NAN if no CV present
- “distToPV” – Shortest distance to any portal vein pixel (in micron); NAN if no PV present
- “#subSegments” – Number of first order branches comprising second order branch
- “comprisingFirstOrderBranches” - Lists all comprising first order branches (ids)
- “partitioning” – Lists for each first order branch subsegment (same order as in ids in previous column) it’s portion of the total second order branch length.

### Network Output file 5

Contains third order branch-specific parameters.
First line is header. One line per third order branch.

Category explanations:

- “dataSet” – Dataset name in which third order branch was found
- “subGraph” - Id of sub-graph in which third order branch was found
- “id” - Unique id of third order branch
- “connectedBranches” - Enumeration of connected third order branches (unique ids)
- “length” – Length of third order branch (in micron)
- “inscrRadius” – Average maximal inscribed sphere radius of network subsegment (measured over maximal inscribed sphere radii of all branch comprising nodes) (in micron)
- “bestFitRadius” – Average best fit radius of network subsegment (measured over best fit radii of all branch comprising nodes) (in micron)
- “distToCV” – Shortest distance to any central vein pixel (in micron); NAN if no CV present
- “distToPV” – Shortest distance to any portal vein pixel (in micron); NAN if no PV present
- “cvPathThirdOrderDegree” - Third order path degree to central vein, indicates the minimal number of third order branches needed to be traversed to reach a central vein; x.5 values indicate dead end branches; Nan if no CV present
- “pvPathThirdOrderDegree” - Third order path degree to portal vein, indicates the minimal number of third order branches needed to be traversed to reach a portal vein; x.5 values indicate dead end branches; Nan if no PV present

- “cvParentBranch” - Id of parent branch in regards to ordering imposed by *cvPathThirdOrderDegree*; Nan if no CV present or branch is root
- “cvChildBranches” - Enumeration of child third order branches (ids) in regards to ordering imposed by *cvPathThirdOrderDegree*; Nan if no CV or children present
- “pvParentBranch” - Analog to *cvParentBranch*
- “pvChildBranches” - Analog to *cvChildBranches*
- “cvChildAngles” - Angles of branch to its child branches in spherical coordinates (Θ, φ); Separately computed and listed (delimiter “|”) for every branching point where child branches are attached; If only one child branches from branching point, second child is the rest of the branch (if branching point is not branch end point); Per branching point listed branches are ordered by id, with first branch’s φ set to 0; Listing is format: *id 1*: (Θ, φ), *id 2*: (Θ, φ) | *id 3*: (Θ, φ); Nan if no CV or children present
- “pvChildAngles” - Analog to *cvChildAngles*
- “cvLengthToParentRatio” - Branch length / *cvParentBranch* length; Nan if no CV present
- “pvLengthToParentRatio” - Analog to *cvLengthToParentRatio*
- “comprisingFirstOrderBranches” - Lists all comprising first order branches (ids)

## Cell Quantification

### Cell Output file

Contains cell-specific parameters.
First line is header. One line per cell.

Category explanations:

- “dataSet” – Dataset name in which cell was found
- “label” – Unique identifier of cell within dataset
- “centroidX” - Cell nuclei’s physical x-coordinate [micron]
- “centroidY” - Cell nuclei’s physical y-coordinate [micron]
- “centroidZ” - Cell nuclei’s physical z-coordinate [micron]
- “distToCV” – Shortest distance to any central vein pixel (measured from cell centroid) (in micron); NA if no CV present
- “distToPV” – Shortest distance to any portal vein pixel (measured from cell centroid) (in micron); NA if no PV present
- “volume” – Volume of the cell (measured by counting voxels) (in $micron^3$)
- “borderCATotal” – Contact area of cell with dataset border (in $micron^2$)

- “borderCARatio” – Deprecated, not computed anymore
- “cellCATotal” – Contact area of cell with other cells (in micron$^2$)
- “cellCARatio” – Ratio of cell-cell contact area to overall surface that is not in contact with dataset border
- “sinusoidCATotal” – Contact area of cell with sinusoids (in micron$^2$)
- “sinusoidCARatio” – Ratio of cell-sinusoid contact area to overall surface that is not in contact with dataset border
- “bileCATotal” – Contact area of cell with bile (in micron$^2$)
- “bileCARatio” – Ratio of cell-bile contact area to overall surface that is not in contact with dataset border
- “nilCATotal” – Contact area of cell with space not allocated to other cells, sinusoids or bile (in micron$^2$)
- “nilCARatio” – Ratio of cell-void contact area to overall surface that is not in contact with dataset border
- “numNuclei” – Number of nuclei within cell
- “nuc1Vol” – Volume of first nucleus. (in micron$^3$)
- “nuc2Vol” – Volume of second nucleus, if found, otherwise NA. (in micron$^3$)
- “nuc3Vol” – Volume of third nucleus, if found, otherwise NA. (in micron$^3$)
- “nuc4Vol” – Volume of fourth nucleus, if found, otherwise NA. (in micron$^3$)

## Recipe for test dataset

This section provides concrete step-by-step instructions and parameter values to reproduce high-quality segmentations and reconstructions for Test dataset 2 that can be downloaded at https://www.hoehme.com/ or http://msysbio.com/.

The raw data channels included in this dataset are already pre-processed (CLAHE and Crop). Therefore, these steps will not be covered in this section, instead you can use the provided files to directly test the pipeline-based segmentation and analysis functionalities of TiQuant 2.0.

Test dataset 2 encompasses the following files:

- **122_DAPI_clahe_cut.tif, 122_DPPIV_clahe_cut.tif, 122_DMs_clahe_cut.tif:**
    - These are the already preprocessed (converted into 8-bit grayscale, CLAHE applied and cropped to a z-region suitable for segmentation) raw data files needed as input for the different segmentation pipelines.
- **vein_central_bin.tif, vein_central_overlay. tif, vein_portal_bin.tif, vein_portal_overlay.tif:**
    - These files hold the results of the vein segmentation. The files ending with '_bin' are binary masks, necessary as input for other segmentation pipelines. The files ending with '_overlay' are for visualization purposes only. Since the datasets features only a central vein, the portal vein segmentation is empty.
- **nuclei_step6_hepNuclei_bin.tif, nuclei_step6_hepNuclei_overlay.tif:**
    - These two files represent the nuclei segmentation. The file ending with '_bin' is a binary mask, used as input for e.g. the hepatocyte reconstruction pipeline. The file ending with '_overlay' is for visualization of the nuclei segmentation and is not meant as input for further processing.
- **sinus_step5_bin.tif, sinus_step5_overlay.tif, sinus_step5_skeleton.tif:**
    - This group of files represents the results of the sinusoid segmentation. The file ending with '_bin' is a binary mask, which is used as input for e.g. the graph extraction pipeline. The file ending with '_overlay' is for visualization of the sinusoid segmentation only. The file ending with '_skeleton' features the skeleton lines of the sinusoidal network, which is input for the graph extraction pipeline.
- **bile_step6_bin.tif, bile_step6_overlay.tif, bile_step6_skeleton.tif:**
    - This group of files represents the results of the bile canaliculi segmentation, the file use is analogue to the sinusoid files.
- **cellShape_step3_bin.tif, cellShape_step3_overlay.tif:**

- These two files represent the cell shape approximation. The file ending with '_bin' is a binary mask, used as input for the hepatocyte analysis pipeline. The file ending with '_overlay' is for visualization of the segmentation and is not meant as input for further processing.

- **Bile_graphX.txt, sin_graphX.txt:**
  - There are two groups of files: The files beginning with the prefix Bile which represent the bile graph, and the files beginning with the prefix sin representing the sinusoidal graph. In both cases a set of enumerated sub-graphs comprise the network within a dataset. This is due to the fact, that a network in a dataset is not necessarily fully connected. Especially at the borders of a dataset subgraphs might reside, that are not linked to the rest of the graph by a connection within the dataset. Isolated graphs are saved separately for technical reasons.

**Vein segmentation**

*If you don't want to overwrite vein_central_bin.tif, vein_central_overlay. tif, vein_portal_bin.tif, and vein_portal_overlay.tif please move them into a sub-directory.*

1. Select 'Segment Veins' in the Pipeline selection drop-down list.
2. Specify the path to *122_DPPIV_clahe_cut.tif* in the 'DPPIV channel' field and *122_DMs_clahe_cut.tif* in the 'DMs channel' field.
3. Click on 'Seed Point Selection'. After a few seconds a window will show up.
4. Hit the 's'-key on your keyboard, notice how the red text changes now into green, which means you are now able to place seed points. Click one time into the cavity of the vein on the right hand side of the dataset to place a seed point which is shown as a crosshairs. If you are not satisfied with its position you can drag it to another position or you can delete it by hitting the 'DEL'-key on your keyboard. Once you are satisfied with its position hit again the 's'-key to quit the seed point input mode and click on 'Finish CV Seed Point Selection'. Since this dataset contains only one central vein and no portal veins click now on the 'Finish Seed Point Selection'-Button. The window will close and you can see the number of placed seed points under 'Central Vein Seed Points' / 'Portal Vein Seed Points' in the parameter table.
5. Click the 'Start Pipeline'-Button.

**Nuclei segmentation**

*If you don't want to overwrite nuclei_step6_hepNuclei_bin.tif* and *nuclei_step6_hepNuclei_overlay.tif please move them into a sub-directory.*

1. Select 'Segment Nuclei in 60x Datasets' in the Pipeline selection drop-down list.
2. Specify the path to *122_DAPI_clahe_cut.tif* in the 'DAPI channel' field.
3. Set Sample region sizes under '1.2) Binary Threshold' - 'Adaptive Otsu-Threshold' to *Sample region size x = 50, Sample region size y = 50, Sample region size z = 5 and Number samples to 5.*
4. Adjust under '1.3) Inverse Hole Filling' the *Majority threshold to 2.*
5. For '1.4) Cavity Filling' reset the *Radius to 12* and the *Minimal fraction to 0.7.*
6. Adjust under '1.5) Closing' the *Kernel radii for x and y to 4* and the *Kernel radius z to 2.*
7. The Kernel radii under '1.6) Opening' have to be set to *x = 4, y = 4, z = 2.*
8. Set under '1.7) Nuclei separation' *Alpha to 0.16.*
9. Under '1.8) Remove and group objects' set *Biggest non-hepatocyte diameter to 6, Smallest hepatocyte diameter to 5.5 and Roundness of hepatocyte nuclei to 0.75.*
10. Click the 'Start Pipeline'-Button.

**Sinusoid and Bile Canaliculi segmentation**

*If you don't want to overwrite sinus_step5_bin.tif, sinus_step5_overlay.tif, sinus_step5_skeleton.tif and bile_step6_bin.tif, bile_step6_overlay.tif, bile_step6_skeleton.tif please move them into a sub-directory.*

1. Select 'Segment Sinusoids + Bile Canaliculi in 60x Datasets' in the Pipeline selection drop-down list.
2. Specify the paths to *122_DPPIV_clahe_cut.tif* in the 'DPPIV channel' field and *122_DMs_clahe_cut.tif* in the 'DMs channel' field.
3. Set under '1.3) Inverse Hole Filling' the *Majority threshold to 9.*
4. Adjust under '1.4) Cavity Filling' the *Minimal fraction value to 0.69.*
5. Under '1.5) Closing & Opening' reset the *radii values for closing to x = 4, y = 4, z = 2.*
6. Select in '2.2) Binary Threshold' *Manual Threshold as DPPIV channel threshold mode* and set the *DPPIV manual threshold to 75.*
7. Click the 'Start Pipeline'-Button.

**Network graph extraction and analysis**

*If you don't want to overwrite the graph files Bile_graphX.txt and sin_graphX.txt please move them into a sub-directory.*

*This pipeline has several switches, and can perform and combine different tasks: Extracting a graph from a '*_skeleton.tif' image or read an already available graph. Optionally analyse a graph. Visualize a graph. The first section will describe, how to use image data to extract a graph, analyze* *and finally visualize it. The second section will outline, how to read an already available graph and visualize it.*

<u>*Graph extraction and analysis:*</u>

1. Select 'Extract and Analyze Graph' in the Pipeline selection drop-down list.
2. Let's assume, you want to extract and analyse the graphs of the bile canaliculi network. The steps for sinusoidal graph extraction are analogous. Select *'Bile Canaliculi Network'* as 'Network type'
3. Specify the paths to *bile_step6_bin.tif* in the 'Specify network segmentation file' field and *bile_step6_skeleton.tif* in the 'Skeleton image' field.
4. Under 'Network Analysis' select 'With Analysis' as 'Analysis mode'. The 'With Analysis' subsection will now be accessible.
5. Specify the paths to *vein_central_bin.tif* in the 'Specify central vein segmentation file' field and *vein_portal_bin.tif* in the 'Specify portal vein segmentation file' field
6. You can assign a dataset ID under 'Specify data set name'. This identifier will then be used throughout the analysis files.
7. In case you want to visualize the graph after pipeline completion check 'Display final graph' under 'Display Options'.

<u>*Graph visualization:*</u>

1. Select 'Visualize Graph' in the Pipeline selection drop-down list.
2. Specify the path to *Bile_graph0.txt* in the *'Graph file'* field. The reader will automatically discover and read all subsequent *Bile_graph* files, if *'Try to load graph sequence'* is activated.

**Hepatocyte shape reconstruction**

*If you don't want to overwrite cellShape_step3_bin.tif* and *cellShape_step3_overlay.tif please move them into a sub-directory.*

1. Select 'Approximate Cell Shape' in the Pipeline selection drop-down list.
2. Specify the paths to *122_DPPIV_clahe_cut.tif* in the 'DPPIV channel' field, *nuclei_step6_hepNuclei_bin.tif* in the 'Hepatic nuclei segmentation' field, *bile_step6_bin.tif* in the 'Bile segmentation' field, *sinus_step5_bin.tif* in the 'Sinusoid segmentation' field, *vein_central_bin.tif* in the 'Central vein segmentation' field and *vein_portal_bin.tif* in the 'Portal vein segmentation' field.
3. Set under 'Parameter' *Alpha to 1*.
4. Click the 'Start Pipeline'-Button.

**Hepatocyte analysis**

1. Select 'Analyze Cells' in the Pipeline selection drop-down list.
2. Specify the paths to *cellShape_step3_bin.tif* in the 'Cell shape binary image' field, *nuclei_step6_hepNuclei_bin.tif* in the 'Nuclei segmentation' field, *bile_step6_bin.tif* in the 'Bile segmentation' field, *sinus_step5_bin.tif* in the 'Sinusoid segmentation' field, *vein_central_bin.tif* in the 'CV segmentation' field and *vein_portal_bin.tif* in the 'PV segmentation' field.
3. You can assign a 'Dataset ID' which is then used as an identifier in the analysis file.
4. Click the 'Start Pipeline'-Button.
5. After pipeline completion an analysis file of the name 'Cell_Output_file.txt' is created in the parent directory of the directory containing the test dataset.